# An Integrated Deep Learning and Dynamic Programming Method for Predicting Tumor Suppressor Genes, Oncogenes, and Fusion from PDB Structures

N.Anandanadarajah[a], C.H.Chu[ab], R.Loganantharaj[ac]

[a] Bioinformatics Research Lab, The Center for Advanced Computer Studies, University of Louisiana at Lafayette, Lafayette, LA, 70503, USA
[b] Informatics Research Institute, University of Louisiana at Lafayette, Lafayette, LA, 70506, USA
[c] Rasbro Technology LLC, 1619 Whitfield St, Sugar Land, TX, 77479, USA
{nishanth.anandanadarajah1@louisiana.edu, henry.chu@louisiana.edu, logan@rasbrotech.com}

**ABSTRACT:**

Mutations in proto-oncogenes (OG) and the loss of regulatory function of tumor suppression genes (TSG) are the common underlying mechanism for uncontrolled tumor growth. While cancer is a heterogeneous complex of distinct diseases, finding the potentiality of the genes related functionality to OG or TSG through computational studies can help develop drugs that target the disease. This paper proposes a classification method that starts with a preprocessing stage to extract the feature map sets from the input 3D protein structural information. The next stage is a deep convolutional neural network stage (DCNN) that outputs the probability of functional classification of genes. We explored and tested two approaches: in Approach 1, all filtered and cleaned 3D-protein-structures (PDB) are pooled together, whereas in Approach 2, the primary structures and their corresponding PDBs are separated according to the genes' primary structural information. Following the DCNN stage, a dynamic programming-based method is used to determine the final prediction of the primary structures' functionality. We validated our proposed method using the COSMIC online database. For the OG vs TSC classification problem the AUROC of the DCNN stage for Approach 1 and Approach 2 DCNN are 0.978 and 0.765, respectively. The AUROCs of the final genes' primary structure functionality classification for Approach 1 and Approach 2 are 0.989, and 0.879, respectively. For comparison, the current state-of-the-art reported AUROC is 0.924. Our results warrant further study to apply the deep learning models to humans' (GRCh38) genes, for predicting their corresponding probabilities of functionality in the cancer drivers.

1. INTRODUCTION

Protein functional annotation is essential in drug development for the target diseases. However, annotating the function of protein through experiments is expensive and time-consuming [1]. Computer-based protein functional annotations reduce the search space for making efficient experimental annotation [1] and thus they have received increased attention recently. Tools to find the protein functional annotations are improved with various methods, such as prediction by sequence [2]–[10], evolutionary relations [11]–[15], protein-protein interactions [16]–[20], protein structures and structure prediction methods [21]–[25], microarrays [26], and combination of data types [27]–[31]. Further, a number of algorithms have been developed to detect protein functional from a given amino acid sequence data [32]. Generally, studies of protein functional detection target all kinds of functionalities, cancer-related or otherwise. Nevertheless, the sub-category of cancer-related functional detection is especially beneficial to cancer treatment.

In addition to classifying complete protein functional detection, computer-based tools can include building predictive models that are useful for prognosis of cancer, classification of cancer types from data sources such as clinical data, SNP's, gene expressions using traditional machine learning algorithms [33]–[41]. Machine learning algorithms that have been successfully used in this area include decision trees, random forest, artificial neural networks, and support vector machines [42]. Deep learning [43], [44] is an increasingly popular machine learning method and has shown remarkable performances for predicting the specificity of DNA and mRNA binding sites [45], functional classification [46], protein folding pattern [47], and for cancer categorization [48]–[53]. A summary of the importance of deep learning in cancer diagnosis can be found in [54]. Cancer is distinctive from other diseases in that it expresses by having tissue growth in an uncontrolled manner due to a failure in the regular cell cycle process. This makes difficult to find the consistent pattern representing their drivers [42]. However, continuous effort in this area makes the data is collected, then categorized and documented [55]. Cancer genes are categorized based on their respective functionality as: (1) Oncogenes (OG): refers to genes which encode for proteins that enable cell growth and proliferation, (2) Tumor suppressor genes (TSG): refers to genes that encode for proteins that control or inhibit cell growth during cell cycle, and (3) Fusion: a hybrid gene formed from two previously separate genes; this type is most problematic because it may cause cancer formed from OG genes, or make loss in TSG function through fusion. Most of the oncogenes, if not all, are due to mutations happened to proto-oncogenes. Accordingly, the roles played by genes in various types of cancer fall into one of these categories [42]: (i) Oncogene (OG), (ii) tumor suppressor gene (TSG), (iii) fusion, (iv) OG and Fusion, (v) OG and TSG, (vi) OG, TSG, and Fusion, and (vii) TSG and Fusion. Earlier studies reported in [42], [54], [54], [56]–[60] and the studies compared by [56] only focused on the core classes OG and TSG, and did not consider the Fusion class. Reference [56] reported a curated set of 99 HiConf cancer genes (48 TSGs, 51 Oncogene), and they considered the rest of the genes related to cancer as unknown (UK; 22 801UK genes). Genomic data and their variance from the cancer genomic atlas (TCGA), International Cancer Genome Consortium (ICGC), and COSMIC were classified by a random forest model integrating five statistical tests to detect the cancer genes [56]. The classification output is the likelihood of OG and TSG [56] and is binary in nature, such as OG vs TSG+UK, OG vs TSG, etc. Even though studies in [56] and [42] did not consider Fusion as a separate class, the studies [61]–[64] show the importance of the Fusion gene. Thus, using the three-dimensional features to automatic detection and prediction or relative probabilities of either Oncogene or cancer suppressor genes or fusion may improve the cancer treatment.

To our knowledge, Reference [42] was the first to classify OG vs TSG using a convolutional neural network (CNN) by utilizing Protein Data Bank (PDB) structures derived from their three-dimensional features. Building on that work, we propose to classify the functionality of the genes as OG vs TSG vs Fusion by a deep learning model with 2D CNN. Input features of the deep learning model are a set of 24 optimal 2D projections obtained from preprocessed PDBs (3D substructure); and each of the 2D projections' channels are presented by PDB's surface residues proposed biochemical properties (different from those in [42]).

Generally, function of a protein (amino acid sequence) given the primary sequence is obtained by the sequence alignment techniques including multiple sequence alignment (MSA). Since obtaining the 3D-structures(e.g. PDB) of a target protein, hereinafter "target-P," is costly and time consuming, the available isoforms of target-P's functionality can be effectively used to predict functionality of target-P. The effectiveness of isoforms' functionality on target-P's functionality may depends on the aligned (overlapping) portion of sequence between the target-P and its isoform, and the number of isoforms shared (partially/fractionally or fully) that aligned with portions of sequence on the target-P's primary structure. Thus, three dynamic programming-based methods are proposed in this paper to give weights to the predicted functionality of substructures. These weights are used to classify the primary structures based on the functional classification of isoforms from the DCNN model.

The rest of the paper is organized as follows. In Section 2, we present the essential step of data preprocessing, including data cleaning and the feature representation based on 2D projections with assigned biochemical properties that we use in the DCNN model. The proposed DCNN model's architecture and parameter selection are summarised in Section 3. Section 4 elaborates on our experiments in gene classifications, carried out at two levels, such as classifying PDBs and proposed methods for classifing primary structure. We used two different approaches to test the classification levels. In the first approach, we do not consider the primary structure information in setting up the training and test sets of PDB substructures. In the second approach, PDB substructures are selected for training and testing based primary structural information. Finally in Section 5 we report the performance of all these classification and draw our conclusions in Section 7.

## 2. DATA SET PREPROCESSING

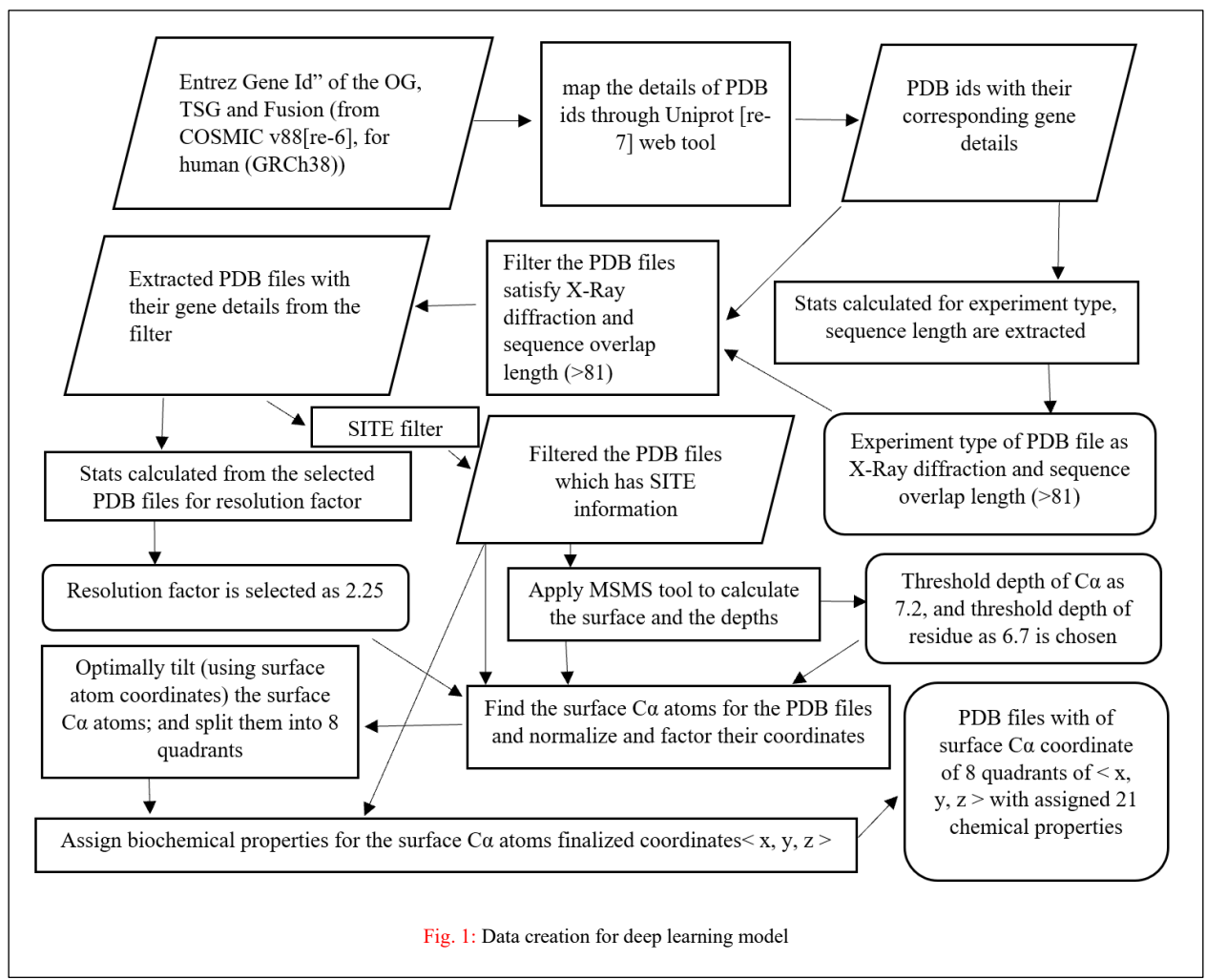

Fig. 1: Data creation for deep learning model

Fig. 1 visualizes the overall data flow of preprocessing. For humans (GRCh38), annotated cancer genes were downloaded from COSMIC v88 [55]. The experiment has focused on using machine learning technique to identify the genes' functional role (probability of OG, TSG, and Fusion) in cancer from their 3D structures. Thus, for preprocessing, the recent version of the COSMIC annotated gene lists' (corresponds to Tier 1) 141 TSG, 80 OG and 94 Fusion genes were first extracted. Then the "Entrez Gene Id" of the OG, TSG, and Fusion (from the census) were used to map the details of PDB ids through UniProt [65]. The mapping has Gene_Symbols with PDB ids and their corresponding, start-end overlap sequence with the gene. This information was used to choose the filtered/selected PDB files (where the PDB files means, the mapped PDB ids; which were filtered with filters such as X-Ray diffraction, and gene overlap sequence length more than the threshold sequence length (81); threshold length selection is reported Appendix 1). The satisfied PDB files were downloaded from protein data bank website [66]. The PDB format exists in standard formats such as macromolecular structure data which achieved by X-ray diffraction, or NMR studies [67]. Some PDB files are cryoelectron microscopy obtained structures [66]. However, majority of PDB files belong to X-Ray defragment (and rapidly increasing by the providers) from the statistics (http://www.rcsb.org/stats/summary and http://www.rcsb.org/stats/growth/growth-xray vs http://www.rcsb.org/stats/growth/growth-nmr vs http://www.rcsb.org/stats/growth/growth-em, last accessed on Sep-26-2020 at 10.52 a.m). Thus, only X-Ray diffraction PDB files were considered in this experiment. Consequently, the genes investigated in this experiment must have at least one PDB file which satisfies all the threshold conditions such as obtained by X-ray diffraction, and overlapping sequence length > 81. The PDB files sequence's overlapping starting-position and end-position were used to calculate the sequence length. If the PDB files are quaternary; then they have more than one polypeptide chain

(sequence start-end positions) of the gene's overall primary structure, then they were left out in the whole experiment.(for more details please check the section 6 "Limitations and assumptions".)

The PDB files satisfying both X-ray diffraction and overlapping sequence length > 81 were selected. Selected PDB files were used to obtain the resolution factor (normalized and factored by resolution factor 2.25, Appendix-2 covers the details of resolution factor selection); the particular nomalization process ensuring that PDB structures with various resolutions could be accommodated without affecting the outcome of the experiment.

From the selected PDB files, only those containing the SITE features were chosen for the analysis. As per [68] SITE information are "*specify residues comprising catalytic, cofactor, anticodon, regulatory or other important sites or environments surrounding ligands present in the structure.*" Thus, these PDB structures may likely play some role in Cancer.

The PDB_ids (where the PDB_id is a PDB file satisfying all X-ray diffraction, overlapping sequence length > 81 and contain SITE feature) were used to train the CNN architecture (mentioned in section 2), and classify the genes (the gene at least has one PDB_id satisfying X-ray diffraction, aligned/ overlapping sequence length > 81 and contain SITE information). These PDB_id's surface Cα atoms features (their coordinates with its chemical property and SITE information) were used to train the CNN architecture (as mentioned in section 2). MSMS tool was used to find the surface Cα atoms; Appendix-3 covers how the threshold depths such as depth of Cα (7.2) and depth of residue (6.7) from the surface, combination depths (less than both threshold depths) used for surface atom selection. The selected surface Cα atoms are optimally tilted and splited as 8 qudrants as mentioned in section 2.1. The biochemical properties were assigned as mentioned in section 2.2.

For the classification of primary structure, the 3-D structures' (PDB_ids') probabilities were weighted by three methods. To do this initially weights were calculated. Section 4, explains the calculation of the weights from their overlap sequence length of PDB_ids(those PDB_ids' should be belong to the corresponding primary structure).

## 2.1 Optimaly tiltation and quadrant separation

These PDB structures can be presented in any orientation; because these PDB structures are uploaded by different authors/ biologists since orentation doesn't matter for them. There may be impact in the performance while the orientation changes for, same PDB structures' surface Cα coordinates presented in different orientation. 3D-convolution is the best way to avoid the classification affected by the orientation. Unfortunately 3D-convolution consumes a lot of memory it is not supported by current technology for this problem. This is because one PDB structures surface Cα coordinates with dimention 256 x 256 x 256 x channels consume around ~2.7 GB for 21 channels, apart from this model's parameter calculations need memory as well.

To get the same performance while if the PDB structures' surface Cα coordinates is presented (rotated) in different orientation, all the PDBs' supposed to have to be titled in their principle direction; so, even if the PDB presented in different orientation, the tiltation make rigid the projections going to be same in all cases. The preservation of scaling through rotating the coordinates is shown in Fig. 2 as Equations (1)-(4)).

---

To do this we first find the covariance matrix of the surface Cα coordinates:

$$X_{(\text{\#of coordinates } \times 3)} = \text{surface C}\alpha \text{ coordinates}$$

$$\text{Cov}(X) = X^T X \quad //\text{find the covariance of the X} \quad (1)$$

Do the eigen value decomposition from the Cov(X) obtained in equation (1), to find the eigen vector of X. Let U, D, V be the singular value decomposition of $X = UDV^T$.

By definition

$$X^T X = V D^2 V^T. \quad (2)$$

Using the eigenvector obtained from Equation (2) we can perform the rotation

$$Y = XV. \quad (3)$$

We can verify that if the rotation preserves the original vector X without affecting the scaling

$$Y^T Y = (XV)^T XV$$
$$= V^T X^T X V$$
$$= V^T (UDV^T)^T (UDV^T) V$$
$$= V^T V D U^T U D V^T V$$
$$= D^2. \quad (4)$$

Equation (4) shows the rotation preserved the scaling.

Fig 2: Rotation of surface Cα coordinates

---

Further, in order to make the models more effectively these tilted PDBs structures' surface Cα coordinates; these surface Cα coordinates are splitted into quadrants (eight), then using the projections (along xy, yz, and xz) of each quadrants separately (128 x 128); thus this data representation endup with 24 projection (8 x3). This make sure save the memory with preserve the positional information for all PDB structures in same way (not dependent on the orientation of PDB structures such it presented in the deposit).

## 2.2 Refining the PDB data files

The PDB_ids are filtered based on primary structural length>81 as shown in Fig.1(Data creation for deep learning model). These filtered PDB_ids may overlap in different classes (like one PDB_id may fall in OG & TSG, or Fusion & TSG or OG & Fusion; or OG & TSG & Fusion). Thus, the mapped PDB_ids needed to be preprocessed and cleaned (to avoid noise to the models such as deep learning model and other) before defining the isoform (protein structure/PDB) to class (OG/TSG/Fusion).

Only the primary structures having higher overlapping PDBs of OG and TSG are removed. From the experiment; the overlapping PDB contained primary structure deltails is shown in Table I; only the Gene_Symbols associates to, OG (PIK3CA corresponding Entrez Gene Id is 5290), and TSG (PIK3R1 corresponding Entrez Gene Id is 5295) were removed. And these primary structures' corresponding group of PDB_ids were also totally removed for training and testing. The dataset of the PDB_ids after this separation is called preprocessed PDB_ids. Those preprocessed PDB_ids are obtained from the remaining primary structures' corresponding group of PDB_ids.

After the intial preprocessing of the PDB_ids, those PDB_ids of any overlaps among OG, TSG, and Fusion classes are also removed. The process ensure that cleaned PDB_ids never have PDB_ids overlapped among OG, TSG, and Fusion classes.

TABLE I: PROBLEMATIC PRIMARY STRUCTURES (CONTAIN OVERLAPPING PDBs) FROM V88 CENSUS USING UNIPORT MAPPING AND PROCESSD

| Class | Gene symbol | Gene_ID | Number of PDBs | | uni-gene(retired) |
|---|---|---|---|---|---|
| | | | group | overlapped | |
| OG | H3F3A | 3020 | 9 | 9 | Hs.726012 |
| | HIST1H3B | 8350 | 39 | 39 | Hs.626666 |
| | PIK3CA | 5290 | 36 | 31 | Hs.732394 |
| | XPO1 | 7514 | 1 | 1 | Hs.370770 |
| TSG | PIK3R1 | 5295 | 38 | 31 | Hs.734132 |
| Fusion | HLA-A | 3105 | 143 | 143 | Hs.713441 |

Since the lack of PDBs to train Fusion, and Fusion class is formed from OG or TSG primary structure; if all the processed PDBs in Fusion's primary structure group overlapped with other classes, then the primary structure group is removed. For an e.x:

ONGO class' Genesymbol PIK3CA, has 36 filtered PDB_ids, among them 31 filtered PDB_ids also appears in TSG; thus these 31 filtered PDB_ids are considered as overlapped.

## 2.3 Atom biochemical properties

The biochemical properties were assigned to 3-D surface Cα coordinates ( $<x, y, z>$ of each quadrants) depend on the corresponding aminoacid. Here the surface Cα coordinate calculation is obtained as mentioned in previous sections. Here different biochemical property ssignment for 3-D surface Cα coordinates is proposed; the normalized properties' two split parts are shown in Table IIA and IIB. Among with that the PDBs' SITE information is considered as the 21'st property. The uploader of PDB file must provide SITE information, whether the surface Cα has SITE (given in PDB file as REMARK 800). Thus, each surface Cα coordinate has 21 property values, such that 20 properties assigned as mentioned in Table IIA and IIB (the last property added dependent on the SITE information; If the surface Cα atom holds SITE, then the feature of SITE assigned as 1 else as 0).

TABLE IIA: PART OF NORMALIZED GENERAL PROPERTIES

| Amino acid | Short | Abbrev. | Side-chain properties | | | | | | | | | |
| --- | --- | --- | --- | --- | --- | --- | --- | --- | --- | --- | --- | --- |
| | | | Hydro-phobic | | pKa§ | Polar | | pH | | Small | | Aromatic or Aliphatic |
| | | | | | | | | acidic (weak acidic 0.3) | basic (weak-0.3; normal-0.6; strong-1) | | | |
| | | | yes | No | | yes | No | | | Yes | No | Aromatic | Aliphatic |
| Alanine | A | Ala | 1 | 0 | 0.191 | 0 | 1 | 0 | 0 | 1 | 0 | 0 | 1 |
| Cysteine | C | Cys | 1 | 0 | 0.156 | 0 | 1 | 1 | 0 | 1 | 0 | 0 | 0 |
| Aspartic acid | D | Asp | 0 | 1 | 0.162 | 1 | 0 | 1 | 0 | 1 | 0 | 0 | 0 |
| Glutamic acid | E | Glu | 0 | 1 | 0.171 | 1 | 0 | 1 | 0 | 0 | 1 | 0 | 0 |
| Phenylalanine | F | Phe | 1 | 0 | 0.179 | 0 | 1 | 0 | 0 | 0 | 1 | 1 | 0 |
| Glycine | G | Gly | 1 | 0 | 0.191 | 0 | 1 | 0 | 0 | 1 | 0 | 0 | 0 |
| Histidine | H | His | 0 | 1 | 0.146 | 1 | 0 | 0 | 0.3 | 0 | 1 | 1 | 0 |
| Isoleucine | I | Ile | 1 | 0 | 0.189 | 0 | 1 | 0 | 0 | 0 | 1 | 0 | 1 |
| Lysine | K | Lys | 0 | 1 | 0.176 | 1 | 0 | 0 | 0.6 | 0 | 1 | 0 | 0 |
| Leucine | L | Leu | 1 | 0 | 0.189 | 0 | 1 | 0 | 0 | 0 | 1 | 0 | 1 |
| Methionine | M | Met | 1 | 0 | 0.173 | 0 | 1 | 0 | 0 | 0 | 1 | 0 | 1 |
| Asparagine | N | Asn | 0 | 1 | 0.174 | 1 | 0 | 0 | 0 | 1 | 0 | 0 | 0 |
| Proline | P | Pro | 1 | 0 | 0.159 | 0 | 1 | 0 | 0 | 1 | 0 | 0 | 0 |
| Glutamine | Q | Gln | 0 | 1 | 0.176 | 1 | 0 | 0 | 0 | 0 | 1 | 0 | 0 |
| Arginine | R | Arg | 0 | 1 | 0.148 | 1 | 0 | 0 | 1 | 0 | 1 | 0 | 0 |
| Serine | S | Ser | 0 | 1 | 0.178 | 1 | 0 | 0 | 0 | 1 | 0 | 0 | 0 |
| Threonine | T | Thr | 0 | 1 | 0.170 | 1 | 0 | 0 | 0 | 1 | 0 | 0 | 0 |
| Selenocysteine | U | Sec | 0 | 1 | 0.155 | 0 | 1 | 1 | 0 | 1 | 0 | 0 | 0 |
| Valine | V | Val | 1 | 0 | 0.194 | 0 | 1 | 0 | 0 | 1 | 0 | 0 | 1 |
| Tryptophan | W | Trp | 1 | 0 | 0.200 | 0 | 1 | 0 | 0 | 0 | 1 | 1 | 0 |
| Tyrosine | Y | Tyr | 0 | 1 | 0.179 | 1 | 0 | 0.3 | 0 | 0 | 1 | 1 | 0 |
| Normalization way explained | | | Not needed binary | | This part only just factored by the (value - min_value) / (max_value-min_value) | Not needed binary | | Between 0-1 so no normalisation needed the 0.3, 0.6 shows the weightage | | Not needed binary | | | |

TABLE IIB: PART OF NORMALIZED GENERAL PROPERTIES

| Amino acid | Short | Abbrev. | General chemical properties | | | | Gene expression and biochemistry | | | Mass spectrometry | |
|---|---|---|---|---|---|---|---|---|---|---|---|
| | | | Avg. mass (Da) | pI | pK1 (α-COOH) | pK2 (α-+NH3) | Essential‡ in humans | | | Mon. mass§ (Da) | Avg. mass (Da) |
| | | | | | | | Yes | No | Conditionally | | |
| Alanine | A | Ala | 0.436 | 0.399 | 0.833 | 0.581 | 0 | 1 | 0 | 0.382 | 0.382 |
| Cysteine | C | Cys | 0.593 | 0.278 | 0.182 | 1.000 | 0 | 0 | 1 | 0.554 | 0.554 |
| Aspartic acid | D | Asp | 0.652 | 0.000 | 0.288 | 0.596 | 0 | 1 | 0 | 0.618 | 0.618 |
| Glutamic acid | E | Glu | 0.720 | 0.038 | 0.455 | 0.379 | 0 | 0 | 1 | 0.693 | 0.693 |
| Phenylalanine | F | Phe | 0.809 | 0.334 | 0.606 | 0.298 | 1 | 0 | 0 | 0.790 | 0.790 |
| Glycine | G | Gly | 0.368 | 0.406 | 0.833 | 0.535 | 0 | 0 | 1 | 0.306 | 0.306 |
| Histidine | H | His | 0.760 | 0.601 | 0.000 | 0.308 | 1 | 0 | 0 | 0.737 | 0.736 |
| Isoleucine | I | Ile | 0.642 | 0.405 | 0.788 | 0.525 | 1 | 0 | 0 | 0.608 | 0.608 |
| Lysine | K | Lys | 0.716 | 0.853 | 0.545 | 0.172 | 1 | 0 | 0 | 0.688 | 0.688 |
| Leucine | L | Leu | 0.642 | 0.399 | 0.803 | 0.515 | 1 | 0 | 0 | 0.608 | 0.608 |
| Methionine | M | Met | 0.731 | 0.365 | 0.500 | 0.283 | 1 | 0 | 0 | 0.704 | 0.705 |
| Asparagine | N | Asn | 0.647 | 0.324 | 0.515 | 0.000 | 0 | 1 | 0 | 0.613 | 0.613 |
| Proline | P | Pro | 0.564 | 0.436 | 0.227 | 0.970 | 0 | 1 | 0 | 0.522 | 0.522 |
| Glutamine | Q | Gln | 0.716 | 0.354 | 0.561 | 0.207 | 0 | 1 | 0 | 0.688 | 0.688 |
| Arginine | R | Arg | 0.853 | 1.000 | 0.030 | 0.136 | 0 | 0 | 1 | 0.839 | 0.839 |
| Serine | S | Ser | 0.515 | 0.358 | 0.591 | 0.247 | 0 | 1 | 0 | 0.468 | 0.468 |
| Threonine | T | Thr | 0.583 | 0.348 | 0.439 | 0.192 | 1 | 0 | 0 | 0.543 | 0.543 |
| Selenocysteine | U | Sec | 0.823 | 0.331 | 0.167 | 0.646 | 0 | 1 | 0 | 0.811 | 0.806 |
| Valine | V | Val | 0.574 | 0.398 | 0.894 | 0.515 | 1 | 0 | 0 | 0.532 | 0.532 |
| Tryptophan | W | Trp | 1.000 | 0.384 | 1.000 | 0.348 | 1 | 0 | 0 | 1.000 | 1.000 |
| Tyrosine | Y | Tyr | 0.887 | 0.353 | 0.606 | 0.247 | 0 | 0 | 1 | 0.876 | 0.876 |
| Normalization way explained | | | This part only just factored by the 1/max_Avg_mass | This part only just factored by the (value - min_value) /(max_value-min_value) | | | Not needed binary | | | This part only just factored by the 1/max_Avg_mass | |

Here the new amino acid "Selenocysteine" now it is represented by "U" is added along with the 20 amino acids. And the new general normalized properties of the 21 amino acids is shown in Table. IIA and Table. IIB (since the number of properties are high for visualization only the table is splitted). One Example (PDB training class ONGO) that has 21'st amino acid.

## 3. DEEP LEARNING MODEL

As mentioned earlier, each PDB file is represented by 21 features associated with the atomic coordinates of eight quadrants with $<x, y, z>$. The approach is almost the same as [42] in the way, such that three-dimensional feature spaces were converted to the 2-D feature maps for each quadrant; inorder to make sure the continuity in the border condition of each quadrants (base zero clearence about 5, is used; such that first quadrant start from -5,-5,-5 for $<x, y, z>$ and so on; further all these quadrants are factored accordingly to make like first quadrant, for an example only the x coordinates of second quadrants is multiplied by -1). Thus, 24 (8x 3=24) independent feature sets associated with three atomic projections on $<x, y>$, $<y, z>$, and $<x, z>$ feature spaces of eight quadrants were generated by conversion from eight$<x,y,z>$features. Therefore, each PDB_id's optimaly tilted 3-D coordinates' each quadrants' was converted to three perpendicular 2-D coordinate spaces. In the next step, each projection was converted to 21 feature maps corresponding to the 21 feature values computed as mentioned in the previous sections.

This approach converts a 3-D structure to 24 feature map sets, with dimensions of $128 \times 128 \times 21$ pixels (21 feature maps of $128 \times 128$). As mentioned in [42] "*Processing the projections is much faster than processing the 3-D structures while not losing information considerably due to the PDB's sparse structure. Furthermore, each feature map set of a projection denotes specific features of the protein while preserving its spatial information.*" Further optimal tiltation and quadrant separation make more efficient. Section 3.1 elaborates, how are these three feature map sets are used for classification of OG Vs TSG Vs Fusion.

## 3.1 CNN Architecture

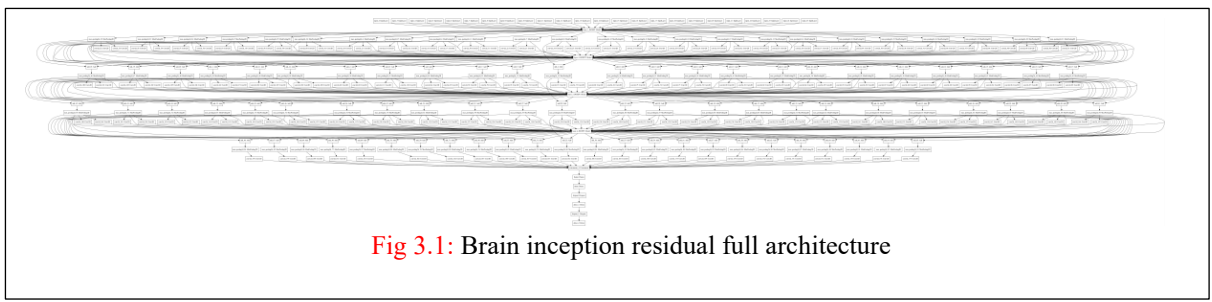

Fig 3.1: Brain inception residual full architecture

The architecture is shown in Fig 3.1 (Expanded figure shown in Appendix-8). The model is named as "Brain-inception-residual"; since the architecture is almost following the brain architecture (because the architecture reuses the same inception feature extractor in different projections; while use different dimensionality reduction networks in each projections to find, if their relative positional information). In order to illustrate the model's overall architecture, feature extraction of only two projections' are shown in Fig 3.5; likewise rest of the 22 projections are going to be followed/fed in visual feature extraction, and then extracted visual features are concatenated, and flatten before feeding in to the Fully connected NN. The model's visual feature extraction inception layers' modules are shown in Fig 3.2 and Fig 3.3. These visual feature extraction layers have naïve inception_V1 [69] in layer_1 as shown in Fig 3.2, inception with dimensionality reduction [69], and residual connection as shown in Fig 3.1 or Fig 3.5. And the architecture's almost all layers are equipped with "Swish" [70] activation function. (Rectified Linear Unit called as "ReLU" is equipped sometimes in the feature extraction layers after Max-pool; Softmax is always equipped in the final of fully connected o/p layer). The final fully connected architecture is shown in Fig 3.4; these experiments used 50% dropout in the fully connected layers to control probably overtraining.

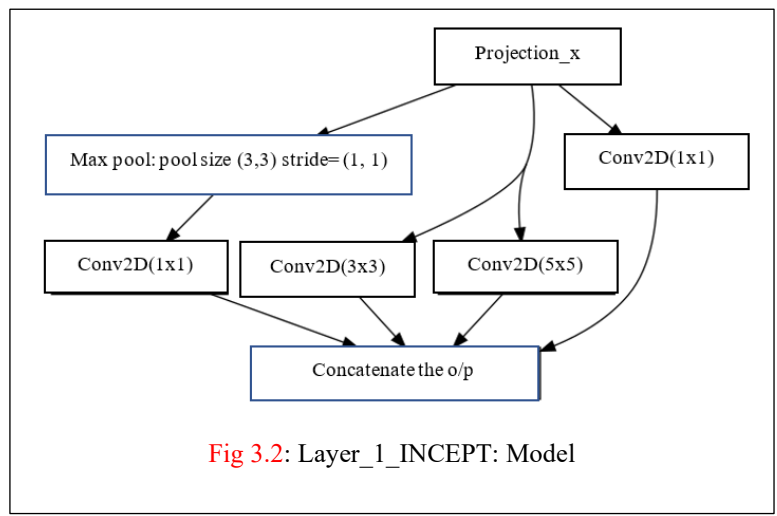

Fig 3.2: Layer_1_INCEPT: Model

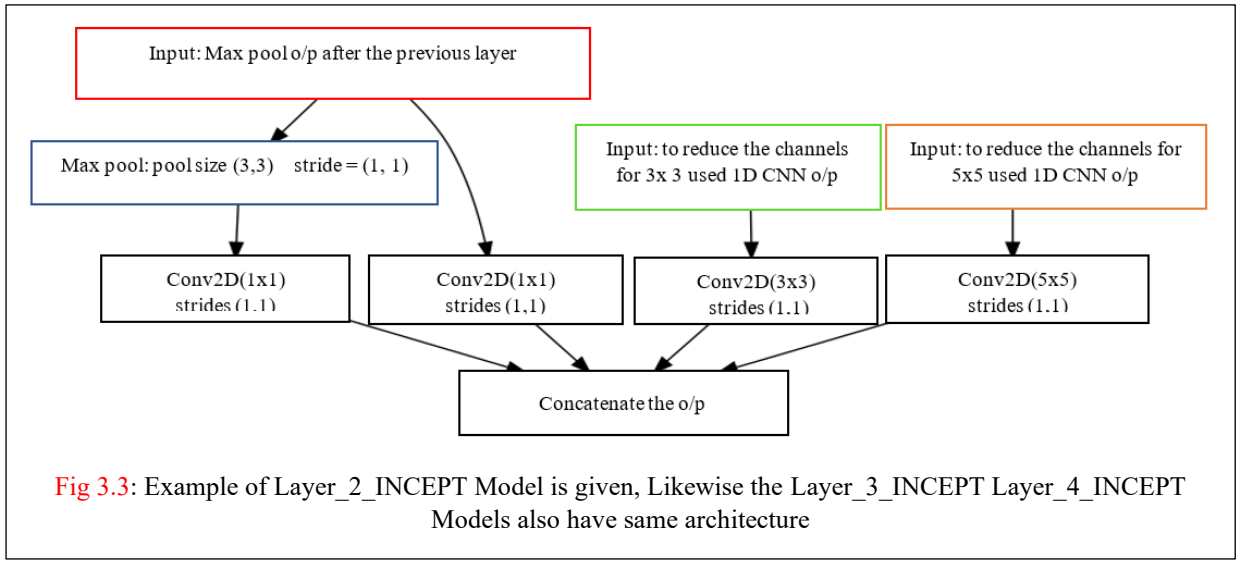

Fig 3.3: Example of Layer_2_INCEPT Model is given, Likewise the Layer_3_INCEPT Layer_4_INCEPT Models also have same architecture

The parameters assigned for these modules (of the brain inception residual model) are mentioned in Table III. The Network section in Table III implies parameters applicable network modules.

Inception modules of layer 1 and layers 2-4 are shown in Fig 3.2 and Fig 3.3 respectively. The number of generated feature maps of inception module's CNN with kernel i x i, is d_CNN_inc_i (therefore d_CNN_inc_1, d_CNN_inc_3, and d_CNN_inc_5 are inception modules' CNN kernels with 1x1, 3x3 and 5x5 respectively). Each inception module's total generated feature map is 256 (d_CNN_inc_1 + d_CNN_inc_3 + d_CNN_inc_5 + d_CNN_1_max). These 256-feature maps are added with the previous layers' Max-Pool's 256-feature maps as shown in Fig 3.5 (the blocks named as "Residual: ADD"). Since the 256-feature maps is expensive to feed in each layers continuously, the layer reducers extract the most wanted feature maps for such as d_lay_1_to_2 (for layer 1 and 2 to end up with 32 features maps) and d_lay_3_to_4 (for layer 3 and 4 end up with 64 features maps); and Max pooling in the layer reducers uses the same pool size for their strides (e.g. in "layer 1 to 2 reducer" or "1->2" uses Max pool with 4x4 filter and 4x4 stride) with valid padding. All the CNNs' and rest of the Max Pools' (excluding layer reducers) have same padding, and their stride size is 1 x 1.

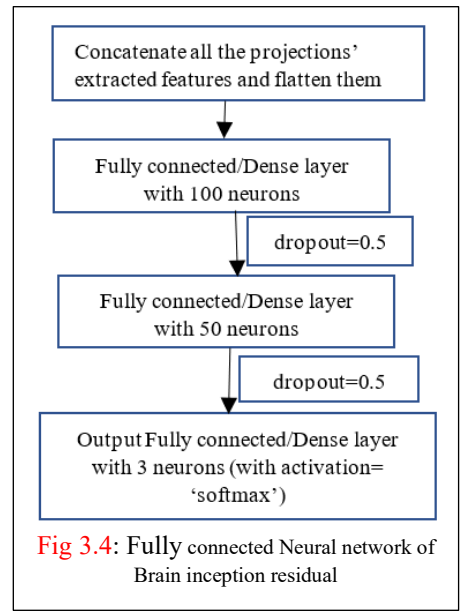

Fig 3.4: Fully connected Neural network of Brain inception residual

From the controlled experiments in base models, the brain-inception-residual model parameters are assigned.

TABLE III: THE BRAIN INCEPTION RESIDUAL DCNN'S PARAMETERS

| Network | Layers | Parameters | Values | Activation | Network | Layers | Parameters | Values | Activation |
|---|---|---|---|---|---|---|---|---|---|
| Inception | 1-4 | d_CNN_inc_1 | 128 | Swish | layer reducer | 1->2 | Max Pool (and stride) | 4x4 | |
| | 1-4 | d_CNN_inc_3 | 64 | Swish | | 2->3 | Max Pool (and stride) | 3x3 | |
| | 1-4 | d_CNN_inc_5 | 32 | Swish | | 3->4 | Max Pool (and stride) | 2x2 | |
| | 1-4 | Max Pool | 3x3 | - | | 4->feature | Max Pool (and stride) | 2x2 | |
| | 1-4 | d_CNN_1_max | 32 | ReLu | | 1,2 | d_lay_1_to_2 | 32 | Swish |
| Dense | 5 | Hidden Neurons | 100 | Swish | | 3,4 | d_lay_3_to_4 | 64 | Swish |
| | 6 | Hidden Neurons | 50 | Swish | Final layer (7) | | Neurons | 3 | Softmax |

The baseline model parameter is mentioned in Appendix-4.

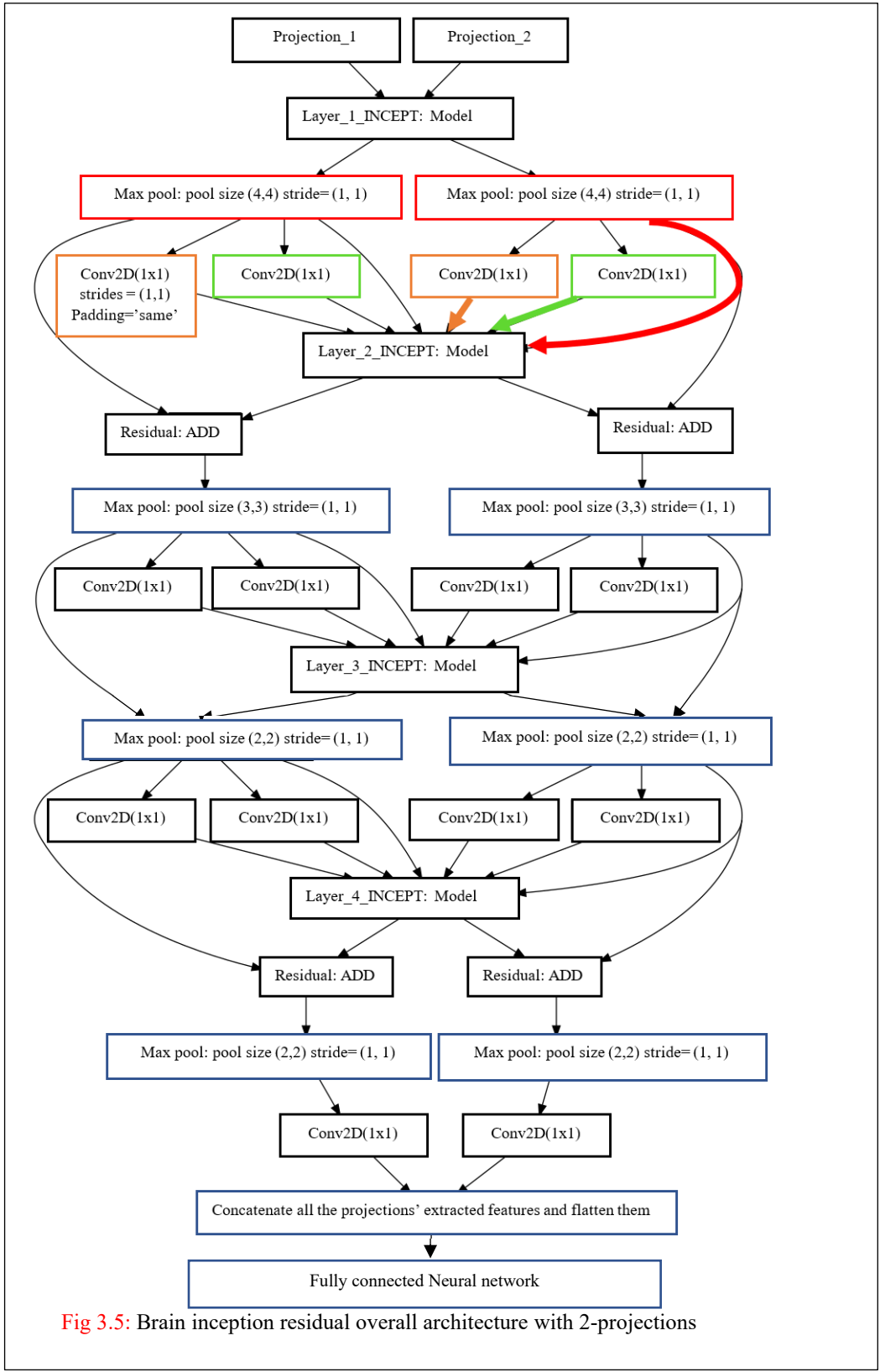

Fig 3.5: Brain inception residual overall architecture with 2-projections

The inception residual convolution/pooling layers extract 24x2x2×64 = 6144 visual features. The model has 2,925,751 trainable parameters (And the parameter calculation is shown in Appendix-5 Table A5.I). For approach 2's total number of train data's feature is and Ratio of Input features/Trainable parameter is 4604. Approach 1's and Approach 2's total number of train data's feature is $1.56 \times 10^{10}$ and $1.34 \times 10^{10}$ respectively; and Ratio of Input features/Trainable parameter is 5,357 and 4604 respectively. (the calultation is shown in Appendix-5)

4. EXPERIMENT

Providing primary structure information while separating PDB structures for training may make a difference in functional detection unless there are lot of PDB structures available. Thus, two approaches proposed to check the impact level of providing primary structure. In approach 1 primary structure wise information is not considered at all, all PDBs are pooled together; on the other hand approach 2 the 3-D structures (PDBs) are separated as training and testing based on primary structural information. Only Tier-1 COSMIC genes are used in both aproaches.

In both approaches; two different functional classifications are done such as functional classification of isoforms and functional classification of primary structure (from the functional classification of isoforms results gained through deep learning model). Hence, in this experiments' data selection is elaborated in section 3.1.

In both approaches training set was only used to train the deep learning model as mentioned in section 3. The classification is elaborated in section 4.2.

In approach 2 the 3-D structures (PDBs) are separated as training and testing based on primary structural information (this make sure same primary structural preprocessed PDB files as mentioned in section 1 not fall in both such as training and testing). [42] just pooled all the PDB files and randomly selected 85% of the data (PDB files) for training and rest for testing; thus it [42] only classify the isoforms (PDB files/3D structures).

## 4.1 Data selection for approach 1 and 2

### 4.1.1 Approach 1

To validate the different combinational (the blend of different primary structures') PDB files' contribution to the functionality among ONGO Vs TSG Vs Fusion, 10-fold cross validation is proposed. Here the PDB files are splitted into 10-folds balanced classes validated. Thus, cleaned (V88 mapping) OGs' 1031, TSGs' 706 and Fusions' 371, are shuffled separately (to randomize the PDB files distribution among each classes), splitted into 10-sets (while splitting makesure each has contains almost same number of PDB files). Please check the Appendix-6 for more details.

### 4.1.2 Approach 2

The PDB files are related to the primary structural information. Therefore, the dataset for training and testing is selected based on the primary structural information. If the primary structure is choosed for training, then all the PDB files corresponding to that primary structure should be in the training set. As mentioned in section 2.2 only cleaned dataset used in training.

The removed/cleaned data consists of 2108 PDB files belongs to 138 genes. From that, 1574 PDB files belong to 105 genes were used for training, and the remaining 33 genes and their corresponding 545 PDB files were used for testing. As mentioned in section 3 (and Appendix-5), the training set's 1574 3D coordinate feature set was converted into three 2D feature sets (1574 x 24=37776). Thus, 37776 with 128 x 128 x 21 feature maps, were used to train the deep learning model. Please check the Appendix-6 for more details about the number of gene (primary structure) and number of PDB files used in the experiment.

## 4.2 Classification

Classification is performed in two ways such as classification of PDB files (Isoforms, as mentioned in section 4.2.1), and classification among primary structure (as mentioned in section 4.2.2). In both ways, five predicted outcomes are presented, such as

1. Class_classified: Presenting the combination of classes (such that first check probabilities of classes higher than 40%, if not check higher than 30%)
2. Most_probable_class: Directly assign the highest probable class such that it assigns mostly one class. Some time two classes are assigned, because two classes share higher probability (like 50% each or 40% each etc.)

And the predicted functional probability among the three classes (ONGO, TSG, and Fusion) are presented.

### 4.2.1 Functional classification of PDBs (Isoforms)

The deep learning model (as mentioned in section 2) is used to predict the preprocessed PDB files (as mentioned in section 1). And the model output the PDB files' probabilities for each class (ONGO, TSG, and Fusion). In approach 1 the classification of PDB files is chosen when the PDB file fell the validation set (among the 10-folds). Classification of rest (Tier 2s' and remaining Tier 1s' gene groups apart from ONGO, TSG, and Fusion) of the COSMIC genes PDB files' are obtained by ensambling the approach 1's best 10-fold models' predictions.

### 4.2.2 Functional classification of primary structure

The primary structure's sequence's specific portion may contribute specific role in functionality. The primary structure's sequence's specific portion's role in functionality can be roughly/ rigidly (based on number of available aligned isoforms and the aligned portion) obtained by integrating the aligned (overlapping) isoforms (PDB structure) functionality and the overlapping sequence information. The overlapping sequence information must account the PDB structure's sequence overlapping length with the corresponding primary structure's sequence and number of other PDB structures shares that overlapping portion. If more than one PDB structures share one portion, then define method/ methods to give weightage to each of these shared PDB structures' functionalities on the primary structures' functionality.

The primary structures' corresponding PDB structures' functional classifications were obtained from section 4.2.1. Sequence information (overlap/aligned starting and ending position of sequence) for each PDBs were gained, while preprocessing (as mentioned section 1). There are plenty of Methods to consider the overlapping (aligned) sequence information to assign the function to the primary structure from their corresponding PDB files' functional probabilities (gained from deep learning model). In this experiment, only 3-Methods are proposed to predict the functional probability of the primary structure (with more than one PDB files with X-Ray defragmentation, overlapping sequence length (>81) and those should have SITE information).

Here onwards all PDB_ids mean PDB files which satisfy X-Ray defragmentation and sequence length (>81) and those should have SITE information).

Methods proposed:

1. Method_1: This method uses the primary structure's all PDB_ids for fraction calculation, chooses the remaining highest length and calculate the fractions.
2. Method_2: uses non-overlapping PDB_ids to cover the primary structure much as possible
3. Method_3: uses overlapping PDB_ids to cover the primary structure much as possible

Finally, the **Ensemble**: This combine (average) all the methods mentioned above (such as Method_1, Method_2, and Method_3). If the primary structure has only one PDB_id then it directly took the probability value of PDB_id, those kinds of primary structures named as Direct. Else primary structure's probability is calculated by using all these three methods.

In these methods, primary structures were considered as groups, where "group" means each primary structure has PDB_ids (group) for the corresponding primary structure. A detailed explanation of three methods are given below.

Further, hypothetical examples for each method is attached as **Appendix-9**, where numerical values are used to make it clear. And **Appendix-10** summarises the complexity of these proposed methods.

### 4.2.2.1 Method_1

The only algorithm analyzing all the PDB_ids in the group (belongs to each primary structure separately). Other methods are not using all PDB_ids in the group. The method extracts the highest length PDB_id for fraction calculation.

---

**Pseudo code for Method_1**
 1> Highest length PDB_id is selected from the left-group.
 2> Then the overlapping fractions (fraction of considered length overlapping with the highest length) of rest of PDB_ids (the group of PDB_ids for the primary structure which overlaps with selected PDB excluding selected PDB_id) with the selected PDB_id are calculated.
 3> From that start and end position of overlaping PDB_ids sequence length is updated (increased accordingly with the overlap with the selected PDB_ids). Leave the highest length PDB_id and the remaining PDB_ids in the group is called as left-group. If the left group has atleast one PDB_id, then use left-group again from step 1.
 4> Finalize overall fractions together, by pooling all the selected highest lengths in step-1. And their overlapping fractions contribution is calculated by normalize them (overlapping fractions) using their summation of selected highest lengths.

---

### 4.2.1.2 Method_2

Uses dynamic programming to cover the maximum length of the primary structure sequence with non-overlapping PDB_ids.

---

**Pseudo code for Method_2**
 1> Starting positions of PDB_ids (for the group) are sorted in ascending order
 2> Iterate until find cov (m)// cover all the PDB_ids ("m" is the total number of PDB_ids in the group)

   cov (0) = 0 //initializing condition for dynamic programming

   if the next PDB_id is not overlapped with the previous one (means starting position of the next PDB_id is higher than the covered PDB_ids.)
     cov(n+1) = cov(n) + length(next_PDB_id)
   if it is overlapped, go back until it non-overlap (example: if, it takes j steps to go back)

$$\text{cov}(n+1) = \max \begin{cases} \text{cov}(n), \\ \text{cov}(n-j) + \text{len}(\text{next\_PDB\_id}) \end{cases}$$

**Exception**: If two or more PDB_ids has same start end position both it considered as one for the calculation

---

### 4.2.1.3 Method_3

The only difference from the Method_2 and Method_3, Method_3 chooses the maximum sequence cover as possible with overlap. Thus negate the overlapped part of jth PDB's primary sequence at j step's back covering.

$$cov(n+1) = \max \begin{cases} cov(n) \\ cov(n-j) + length(next\_PDB) - overlapped\_with\_j\_th\_PDB \end{cases}$$

Here, it only considers partial overlap not full overlap.

Exception: As mentioned in Method_2 if two or more PDB_ids have same start end positions both, are considered as one for the calculation

## 5. Results

### 5.1 Approach 1 results (10-fold cross validation with mixed primary structural PDBs)

**Table IV:** Approach 1's(10-fold cross validation's)Area under the curves for Functional classification of PDBs'(using DCNN model) and gene's Primary structure (using DCNN and Ensamble+direct methods)

| AUROCs | Functional classification of PDBs' | | | | | Functional classification of gene's Primary structure (using Ensemble + direct) | | | | |
|---|---|---|---|---|---|---|---|---|---|---|
| Classification 10_fold | ONGO | TSG | Fusion | Micro average | Macro average | ONGO | TSG | Fusion | Micro average | Macro average |
| ONGO Vs TSG Vs Fusion | 0.974 | 0.976 | 0.961 | 0.977 | 0.971 | 0.981 | 0.952 | 0.834 | 0.942 | 0.925 |
| ONGO Vs TSG | 0.978 | 0.978 | N/A | 0.979 | 0.978 | 0.989 | 0.989 | N/A | 0.989 | 0.990 |
| ONGO Vs Fusion | 0.975 | N/A | 0.975 | 0.986 | 0.975 | 0.919 | N/A | 0.919 | 0.925 | 0.926 |
| TSG Vs Fusion | N/A | 0.981 | 0.981 | 0.986 | 0.981 | N/A | 0.863 | 0.863 | 0.912 | 0.876 |

#### 5.1.1 Approach 1's functional classification of PDBs (Isoforms) results

The PDB wise classification of Approach 1's confusion matrix is shown in Fig 5.1.1. Functional 10-fold cross validation accuracies of the following classes ONGO, TSG, Fusion and all combined, are 94.67%, 94.33%, 88.68% and 93.5% respectively.

| valid | ONGO | TSG | Fusion |
|---|---|---|---|
| ONGO | 976 | 32 | 22 |
| TSG | 45 | 666 | 20 |
| Fusion | 10 | 8 | 329 |

**Fig 5.1.1:** Confusion matrix of Approach 1 classification of PDBs in 10-fold crossvalidation on ONGO Vs TSG Vs Fusion

Only ONGO vs TSG is calculated by normalizing the ONGO and TSG probabilities while neglecting the Fusion class (Note, if one PDB classified as Fusion class only then corresponding PDB's probability is given to ONGO and TSG as 0.5 to each); which is shown in Fig 5.1.2 . Likewise, binary classification among ONGO Vs Fusion and TSG Vs Fusion also performed; and shown in Fig 5.1.3 and Fig 5.1.4 respectively. Where, over all accuracies of

ONGO vs TSG: 93.64 %, ONGO vs Fusion: 94.07 %, and TSG vs Fusion: 94.41 % respectively.

ROCs of the classification of PDBs are shown Fig 5.1.5 to Fig 5.1.8, with their corresponding AUROC. Further, the summery of the AUROCs is mentioned in Table IV; where all the AUROCs are above 0.96 and, binary classification of ONGO Vs TSG's AUROC is .978.

| | Given | |
|---|---|---|
| valid | ONGO | TSG |
| ONGO | 985 | 35 |
| TSG | 46 | 671 |

**Fig 5.1.2:** Confusion matrix of Approach 1 classification of PDBs in 10-fold crossvalidation on ONGO Vs TSG

| valid | ONGO | Fusion |
|---|---|---|
| ONGO | 1003 | 35 |
| Fusion | 28 | 336 |

**Fig 5.1.3:** Confusion matrix of Approach 1 classification of PDBs in 10-fold crossvalidation on ONGO Vs Fusion

| valid | TSG | Fusion |
|---|---|---|
| TSG | 683 | 38 |
| Fusion | 23 | 333 |

**Fig 5.1.4:** Confusion matrix of Approach 1 classification of PDBs in 10-fold crossvalidation on TSG Vs Fusion

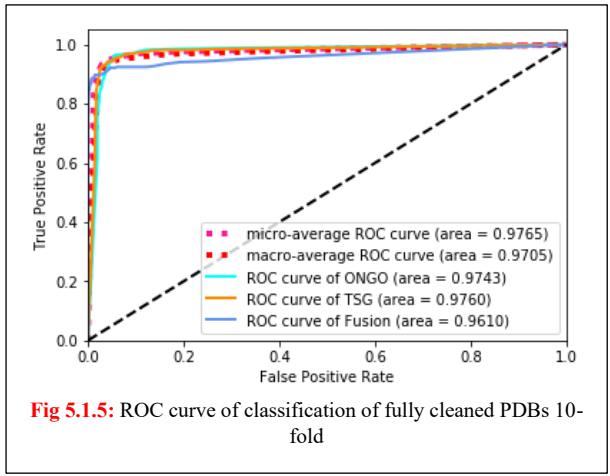

**Fig 5.1.5:** ROC curve of classification of fully cleaned PDBs 10-fold

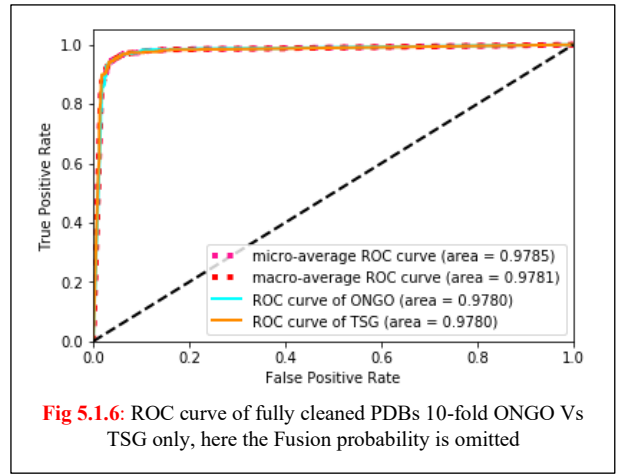

**Fig 5.1.6:** ROC curve of fully cleaned PDBs 10-fold ONGO Vs TSG only, here the Fusion probability is omitted

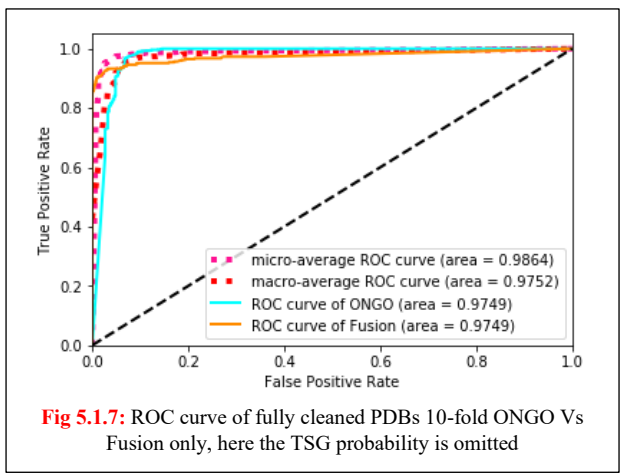

**Fig 5.1.7:** ROC curve of fully cleaned PDBs 10-fold ONGO Vs Fusion only, here the TSG probability is omitted

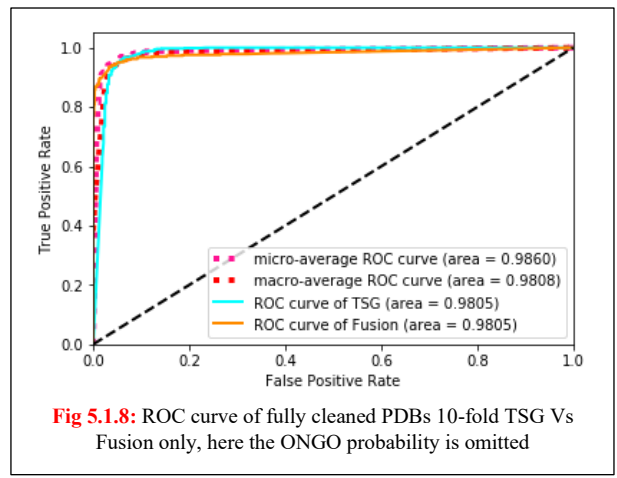

**Fig 5.1.8:** ROC curve of fully cleaned PDBs 10-fold TSG Vs Fusion only, here the ONGO probability is omitted

### 5.1.2 Approach 1's functional classification of gene's primary structure (using Ensemble + direct)

| valid | ONGO | TSG | Fusion |
|---|---|---|---|
| ONGO | 43 | 3 | 3 |
| TSG | 2 | 60 | 7 |
| Fusion | 0 | 1 | 19 |

**Fig 5.1.9:** Confusion matrix of Approach 1 classification of primary structures with 10-fold crossvalidation and methods (from Ensemble and Direct) on ONGO Vs TSG Vs Fusion

| valid | ONGO | TSG |
|---|---|---|
| ONGO | 43 | 4 |
| TSG | 2 | 60 |

**Fig 5.1.10:** Confusion matrix of Approach 1 classification of primary structures with 10-fold crossvalidation and methods (from Ensemble and Direct) on ONGO Vs TSG

| valid | ONGO | Fusion |
|---|---|---|
| ONGO | 45 | 9 |
| Fusion | 0 | 20 |

**Fig 5.1.11:** Confusion matrix of Approach 1 classification of primary structures with 10-fold crossvalidation and methods (from Ensemble and Direct) on ONGO Vs Fusion

| valid | TSG | Fusion |
|---|---|---|
| TSG | 63 | 9 |
| Fusion | 1 | 20 |

**Fig 5.1.12:** Confusion matrix of Approach 1 classification of primary structures with 10-fold crossvalidation and methods (from Ensemble and Direct) on TSG Vs Fusion

Genes' primary structure wise functional classification (using an ensemble method and direct) of Approach 1's confusion matrix is shown in Fig 5.1.9. ONGO, TSG, Fusion and all combined 10-fold cross validation accuracies obtained from functional classification of primary structures' (Genes'), are 95.56%, 93.75% and 65.52% and 88.41% respectively.

Only ONGO vs TSG is calculated by normalizing the ONGO and TSG probabilities while neglecting the Fusion class (same probability calculation as mentioned previous section). Binary classifications' confusion matrices are shown in Fig 5.1.10, Fig 5.1.11 and Fig 5.1.12. Where, binary classifiers' over all accuracies are

ONGO vs TSG: 94.5%   ONGO vs Fusion: 87.8% and    TSG vs Fusion: 89.24 % respectively.

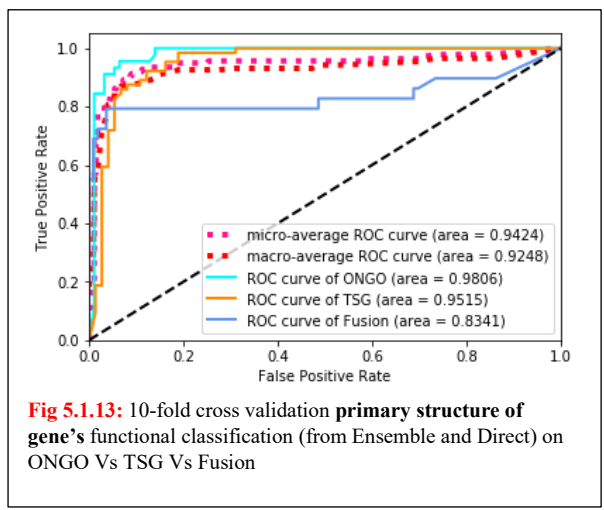

**Fig 5.1.13:** 10-fold cross validation **primary structure of gene's** functional classification (from Ensemble and Direct) on ONGO Vs TSG Vs Fusion

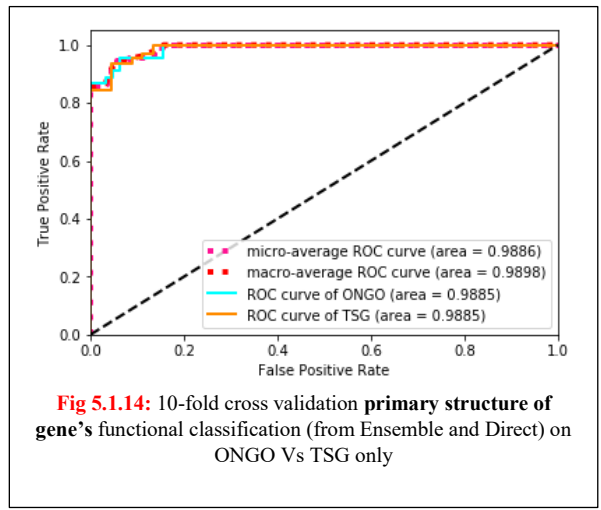

**Fig 5.1.14:** 10-fold cross validation **primary structure of gene's** functional classification (from Ensemble and Direct) on ONGO Vs TSG only

ROCs of the Genes' primary structure wise functional classification are shown Fig 5.1.13 and Fig 5.1.14, with their corresponding AUROC. Further, the summery of the AUROCs is mentioned in Table IV; where, only ONGO Vs TSG classifiers' AUROC is .988.

### 5.2 Approach 2 results (the primary structural genes with their corresponding PDBs are separated as Train and test sets depending on the genes' primary structural information)

#### 5.2.1 Approach 2's Functional classification of PDBs (Isoforms) results

|   |   | Given |   |
|---|---|---|---|
| Test | OG | TSG | Fusion |
| OG | 326 | 20 | 6 |
| TSG | 119 | 59 | 12 |
| Fusion | 0 | 0 | 0 |

**Fig 5.2.1:** Confusion matrix of Approach 2 classification of testing set PDBs on ONGO Vs TSG Vs Fusion

| valid | OG | TSG | Fusion |
|---|---|---|---|
| OG | 83 | 1 | 2 |
| TSG | 5 | 87 | 12 |
| Fusion | 0 | 0 | 12 |

**Fig 5.2.2:** Confusion matrix of Approach 2 classification of validation set PDBs on ONGO Vs TSG Vs Fusion

| Train | OG | TSG | Fusion |
|---|---|---|---|
| OG | 486 | 9 | 110 |
| TSG | 15 | 535 | 77 |
| Fusion | 0 | 0 | 140 |

**Fig 5.2.3:** Confusion matrix of Approach 2 classification of training set PDBs on ONGO Vs TSG Vs Fusion

The PDB wise classification of Approach 2's Test set confusion matrix is shown in Fig 5.2.1. From that none of the test set of PDBs are distinguished by the DCNN (classifier of Approach 2) as Fusion class. Further, the PDB wise classification of Approach 2's validation and Train set confusion matrixes are shown in Fig 5.2.2 and Fig 5.2.3 respectively.

But the DCNN (classifier of Approach 2) have some capability to distinguish between ONGO and TSG. Binary classification of ONGO vs TSG (following the same probability normalization as mentioned in 3.1.2.1.1"Approach 1's functional classification of PDBs (Isoforms) results") for the test set, and validation set got 73.47%, and 96.59% respectively. The accuracy of 73.47% give a validity of informational gain in ONGO Vs TSG even primary structural gene wise separation presented in Train and Test set (some primary structures in Trian set and Test set has same kind of functionality). If the number of PDBs increased, then the classifiers may perform better in functionally classifying ONGO and TSG, even if they separated by primary structure wise.

However, as expected it has some capability of separate Fusion class in validation set (Binary classification of ONGO vs Fusion, and TSG vs Fusion got over all accuracy as 84.21% and 87.72% accordingly). The possible reason behind the lack of performance comparable to the Approach 1; is due to lack of number of PDBs to train the DCNN (since considerable amount of PDBs were allocated to test; such as ~24% primary structural genes were used in Test set, that contains ~34% of PDBs in whole filtered and cleaned PDB set) of Approach 2.

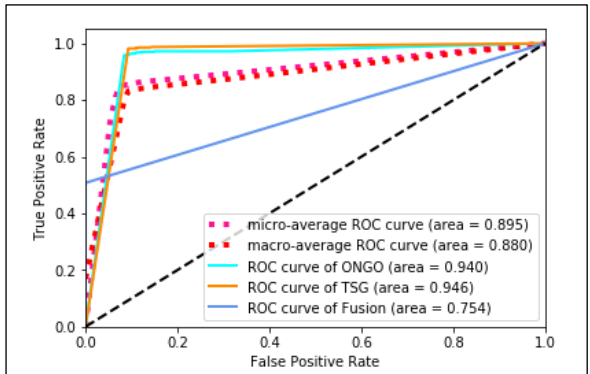

Fig.5.2.4: Approach 2's ROC curve of fully cleaned PDBs **train** set ONGO Vs TSG Vs Fusion

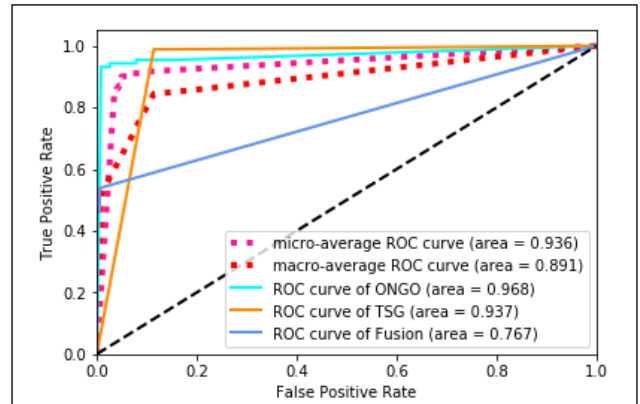

Fig.5.2.5: Approach 2's ROC curve of fully cleaned PDBs **valid** set ONGO Vs TSG Vs Fusion

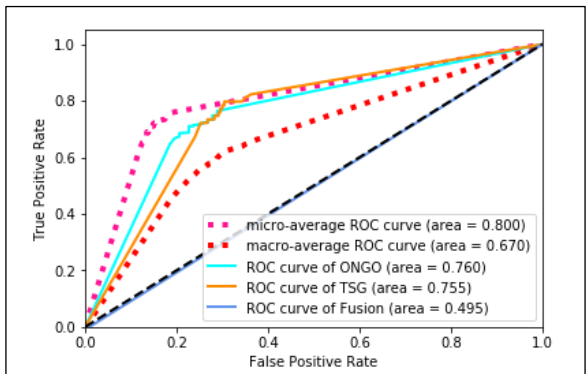

Fig 5.2.6: Approach 2's ROC curve of fully cleaned PDBs **test** set ONGO Vs TSG Vs Fusion

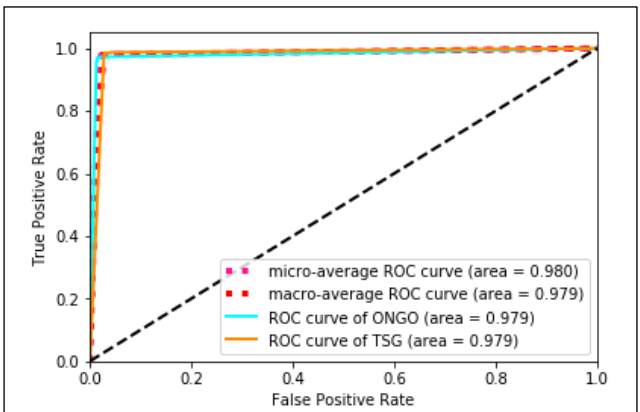

Fig 5.2.7: Approach 2's ROC curve of fully cleaned PDBs **train** set ONGO Vs TSG only, here the Fusion probability is omitted

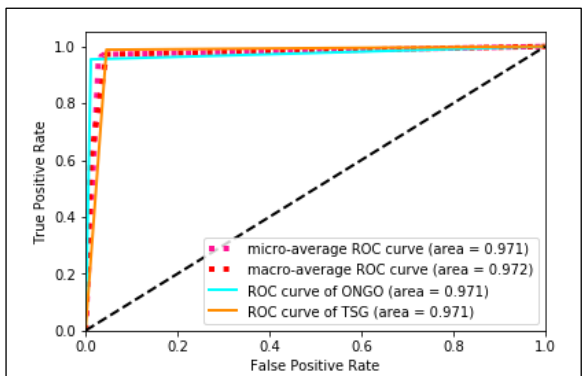

Fig 5.2.8: Approach 2's ROC curve of fully cleaned PDBs **valid** set ONGO Vs TSG only, here the Fusion probability is omitted

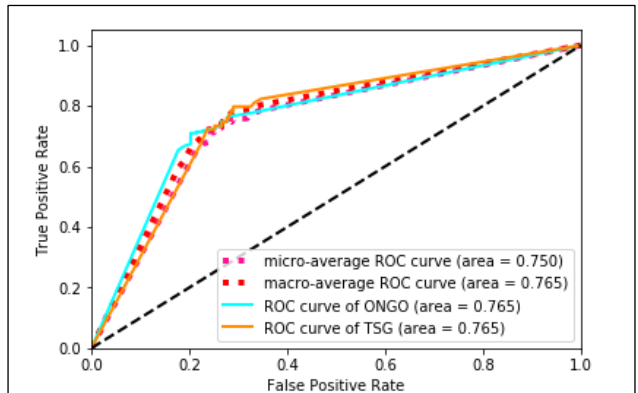

Fig 5.2.9: Approach 2's ROC curve of fully cleaned PDBs **test** set ONGO Vs TSG only, here the Fusion probability is omitted

Classification of PDBs' ROCs are shown in Fig 5.2.4 to Fig 5.2.9, with their corresponding AUROC. Fig 5.2.4 to Fig.5.2.6 represents ONGO Vs TSG Vs Fusion, for Train, Valid and Test sets accordingly. Fig 5.2.7 to Fig 5.2.9 represents binary classification of ONGO Vs TSG, for Train, Valid and Test sets accordingly.

### 5.2.2  Approach 2's functional classification of gene's Primary structure (using Ensemble + direct)

| Test | Given | | |
|---|---|---|---|
| | OG | TSG | Fusion |
| OG | 7 | 1 | 2 |
| TSG | 6 | 13 | 4 |
| Fusion | 0 | 0 | 0 |

**Fig 5.2.10**: Confusion matrix of Approach 2 classification on Test sets' primary structures with DCNN(obtained by train set) and methods(from Ensemble and Direct)

| Train | Given | | |
|---|---|---|---|
| | OG | TSG | Fusion |
| OG | 31 | 0 | 7 |
| TSG | 1 | 50 | 13 |
| Fusion | 0 | 0 | 3 |

**Fig 5.2.11**: Confusion matrix of Approach 2 classification on Train sets' primary structures with DCNN(obtained by train set) and methods(from Ensemble and Direct)

Since none of the PDBs of test set are classified as Fusion class by the DCNN (as shown in Fig. 5.2.1), the methods not going to predict the Fusion class in Test set at all. But the ONGO Vs TSG performance can be evaluted for the primary structural Function prediction as shown in Fig.5.2.10 (since no difference in confusion matrixes of binary classification; the 3-class confusion matrixes are only presented). Further, Fig.5.2.11 shows the Training sets primary structural Function predictions; here the combined Train and Validation PDBs classified functional details are used. The ONGO vs TSG vs Fusion primary structural function prediction of Test sets' ROCs are shown in Fig 5.2.12; further binary classification of ONGO vs TSG is shown in Fig.5.2.13.

From confusion matrix of Test set (Fig 5.2.1), the capability of distinguish between classes is very low, the classifier is almost biased to TSG, due to the lack of PDBs in TSG compare to ONGO, in training data creation higher priority is given to TSG (if not, it does not classify TSG at all; due to biased of higher number of ONGO PDBs; ~28% primary structures of ONGO placed in Test set). If the number of the primary structures of TSG incresed, then the classifier may perform better. Even in the training set's Fusion class classififction is very low, but the Fusion class classififcation performance is better in Approach 1 (10-fold crossvalidation); this validates, if more PDBs are there then the Approach 2 can applicable to check the performance. As per the stats in PDB databank in Future Approach 2 may applicable.

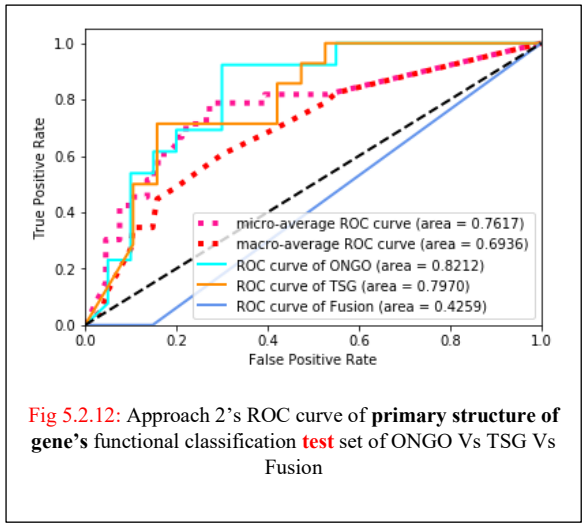

Fig 5.2.12: Approach 2's ROC curve of **primary structure of gene's** functional classification **test** set of ONGO Vs TSG Vs Fusion

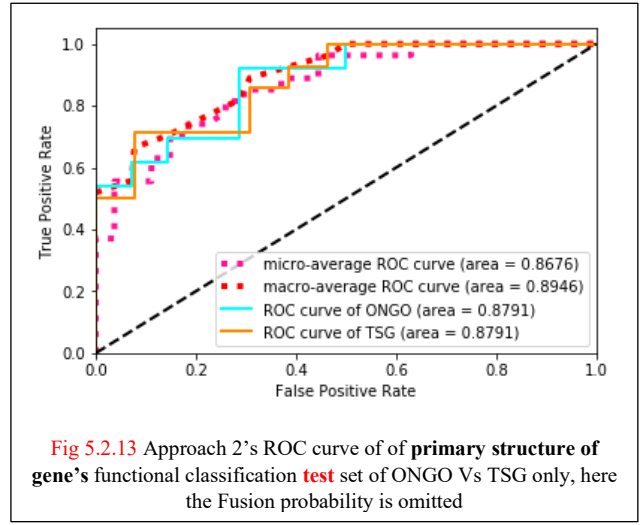

Fig 5.2.13 Approach 2's ROC curve of of **primary structure of gene's** functional classification **test** set of ONGO Vs TSG only, here the Fusion probability is omitted

### 5.3 Comparison with the available state-of-the-art methods for OG vs TSG identification

Since already developed methods [42] and [56] were focused on classifing OG vs TSG or OG vs TSG vs Unknown; In all these methods, Fusion class is not considered at all . Thus Table.XI compares the AUROC values reported in this study with the AUROC values reported by the state-of-art statistical methods [56], [71], [72] for OG vs TSG identification, and the [42].

From Table V, DCNN applied to the PDB structure, performance in this study is far better than [42]; since this study only considering the PDB structure's SITE information. The approach 1's outperforms (since minimum AUROC of approach 1 is 0.94) all (maximumAUROC among other is 0.924) the other state-of-art statistical methods' OG vs TSG identification, and the [42]. Even the approach 2's lowest AUROC is 0.754; because the trainset test set split based on primary structure reduce the number of PDBs to train as well as test; this would be overcome in the future by growing PDB deposits.

**TABLE V:** AUROC OF OG/TSG IDENTIFICATION USING THE STATISTICAL METHODS REPORTED BY [42], [56] AND OUR MODEL.

| Method | | | Description | AUROC |
|---|---|---|---|---|
| Truncation Rate | | | of truncating events* | 0.922 |
| Unaffected Residues | | | Intra-gene mutation clustering/recurrence* | 0.479 |
| VEST Mean | | | Functional impact bias* | 0.710 |
| Patient Distribution | | | Bias in patient labels* | 0.556 |
| Cancer Type Distribution | | | Cancer type bias* | 0.612 |
| Oncodrive-fm | | | Gonzalez-Perez and Lopez-Bigas [72] | 0.725 |
| OncodriveCLUST | | | Tamborero et al., [71] | 0.597 |
| Random Forest | | | Ensemble on the first 5 methods (*) | 0.924 |
| [2] Model's PDB vise functional classification | | | DCNN applied to the PDB structure | 0.887 |
| Proposed PDB vise functional classification | | | Using Approach1's DCNN's probability | 0.978 |
| | | | Using Approach2's DCNN's probability | 0.765 |
| Proposed primary structure vise Functional classification of genes | Dynamic programming methods | Method_1** | Using Approach1's DCNN's probability | 0.997 |
| | | | Using Approach2's DCNN's probability | 0.871 |
| | | Method_2** | Using Approach1's DCNN's probability | 0.94 |
| | | | Using Approach2's DCNN's probability | 0.754 |
| | | Method_3** | Using Approach1's DCNN's probability | 0.979 |
| | | | Using Approach2's DCNN's probability | 0.761 |
| | | Ensemble on the last 3 methods (**) | Using Approach1's DCNN's probability | 0.99 |
| | | | Using Approach2's DCNN's probability | 0.833 |
| | (using Ensemble + direct) | | Using Approach1's DCNN's probability | 0.989 |
| | | | Using Approach2's DCNN's probability | 0.879 |

Using Approach1's DCNN's probability: Means Approach-1's 10-folds DCNN validation on PDB structures' probabilities are used by the methods to functionally classify the primary structures.
Using Approach2's DCNN's probability: Means Approach-2's test sets' primary structures' functionally classification based on the DCNN (trained by approach-2's training set) classification on PDB structures' probabilities.

Further Tier_1 and Tier_2 overall classification is attached in link (https://drive.google.com/drive/folders/1YCuMVPhAy7tIdGFiebInZ_6-kBAH1kMM?usp=sharing).

Inorder to show the results (in the link) representation by the model and methods. Please check the Appendix-7

# 6. Limitations and assumptions

Mapping from gene to protein structures (examined details dataprocessing section). First, these genes were mapped into PDB_ids (to retrieve isoforms). Since one gene may has different isoforms (PDB_ids). Thus, main question is, "how to use(differentiate/weightage) isoforms' functional contribution to obtain the gene's functionality?". One approach for this question is use the isoform's overlapping sequence length (primary structure and where it overlaps?) with the corresponding gene. In order to retrieve the functionally effective isoforms, that are obtained/decoded from the gene (to maintain the gene's functionality), first all the PDB_ids' (Isoforms') sequence length (primary structures) were pooled; and the threshold (80) was chosen as mentioned in Appendix-1. Thus, all the PDB_ids' (Isoforms') corresponding genes' sequence length higher than threshold (80) were chosen.

These PDB_ids may overlap in different classes (like one PDB_id may fall in ONCO & TSG, or Fusion & TSG or ONCO & Fusion). Thus, the mapped PDB_ids needed to be preprocessed (to avoid noise to the models such as deep learning model and other) before defining the isoform (protein structure/PDB) to class (ONCO/TSG/Fusion). In the preprocessing step, the overlapped- PDB_ids (PDB_ids' corresponding genes' class overlap) is tackled by the PDB_ids overlapped in all classes are left/removed (this doesn't mean, those PDB_ids are not contributing to the functionality). So, the preprocessed PDB_ids never have PDB_ids overlapped in classes ONCO & TSG, nor Fusion & TSG nor ONCO & Fusion.

PDB structures are created in different ways: Such as X-ray diffraction, NMR etc [67]. From the stats (select the 'overall'/ 'by X-ray'/'by NMR'/etc in selections under "Growth of Released Structures Per Year" at http://www.rcsb.org/stats/ on 11-02-2019 at 6.14pm) most and rapidly growing, deposits are X-ray diffraction; so in these projects only those structures (structures obtained by X-ray diffraction) are considered.

PDB structures are in different resolutions (most of them fell between 1.0 to 4.6, like 1.0<resolution<4.6;as per stats http://www.rcsb.org/stats/distribution_resolution on 11-02-2019 at 6.46pm). So, it is normalized and factored (by 2.25); the detailed stats of factor selection, is elaborated on Appendix-2 ("Resolution factor selection for normalization").

Some genes are not mappable to PDB_ids(to retrieve isoforms/PDBs). For an example in Tier_2, gene with Entrez GeneId (378938) is the only one gene belong to ONGO_TSG_Fusion class, but the gene is not mappable by UNIPROT. So, I left that gene (Entrez GeneId:378938). This doesn't mean the gene correspond to Entrez GeneId (378938), not contributes to the Cancer; since the lack of mapping capability available now (due to the lack of available literatures/works have done so far on these kind of genes), the methods proposed here are not applicable for these kind of genes.

Some PDB_ids contain more than one chain (primary structure sequence). Those, PDB ids were handled in two ways depend on the chain information. Such as,

- if the PDB has the same sequence (polypeptide) chains for all the chains it has, then it is considered in the experiment.

    E.x:    3UD1; X-ray; 2.00 A; A/B/C=911-1233

    This 3UD1 contains three chains as A, B, and C and all are same

    1K9I; X-ray; 2.50 A; A/B/C/D/E/F/G/H/I/J=250-404.

    These PDBs are included in the functional detection. Since the methods, used/proposed in the paper can use the sequence information to predict the gene class from the predicted class(using the deep learning model) of PDB.

- If the PDB has at least one different sequence (polypeptide) chain among all the chains, it has. Then, it is left.

    Ex:    5X83; X-ray; 3.00 A; A/C=721-815, B/D=844-1043.

    Such this contain two different chains that's why it is left. The methods used/proposed in the paper to predict functional probability of the given gene, is obtained using the sequence information (overlap information), to give weightage to the PDB's probabilities (predicted by deep learning model). Thus, if the PDB has different sequence (polypeptide) chain, then the weightage can't be define to that PDB's functional probability (predicted by deep learning model), by the methods proposed in the paper (Since, the functionality of the corresponding PDB comes from either one of the sequence (polypeptide) chains or group of the chains; as per my knowledge, there is no known way to define which peptide chain's surface residue provide the functionality). Further, the number of structures of these kinds is very low, thus including these kinds of PDBs in the experiment may lead to an unwanted noise or make complications to obtain the gene's functionality from the 3D structures. This doesn't mean, this kind of PDBs not contributing to the functionality.

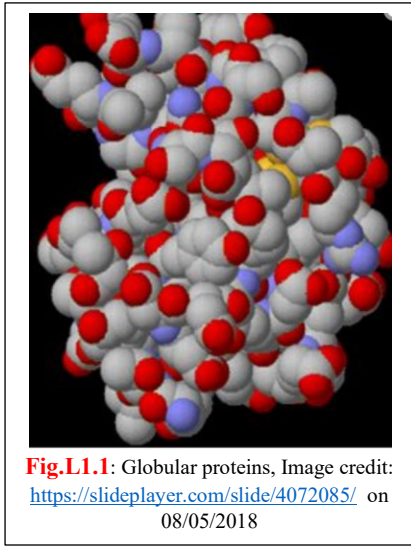

**Fig.L1.1**: Globular proteins, Image credit: https://slideplayer.com/slide/4072085/ on 08/05/2018

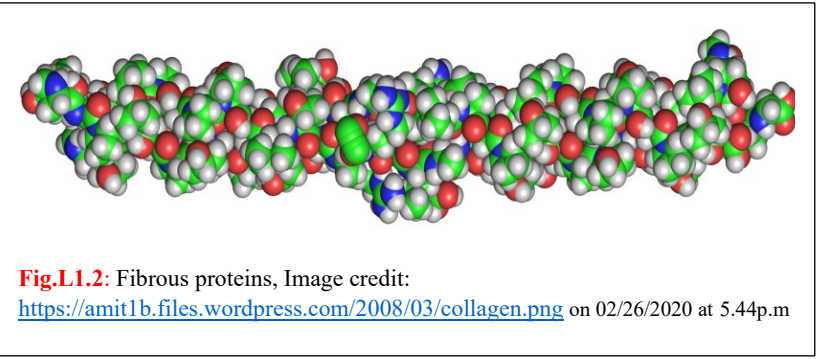

**Fig.L1.2**: Fibrous proteins, Image credit: https://amit1b.files.wordpress.com/2008/03/collagen.png on 02/26/2020 at 5.44p.m

Defining the surface of protein structure is problematic; due to their variety of shapes as shown in Fig.L1.1 and Fig.L1.2. This is solved by using MSMS tool, detailed description of surface detection is mentioned in the Appendix-3 (in the project surface amino acid locations is found, using their corresponding Cα backbone detection).

## 6.1 Limitaion of Database (COSMIC)

The census data used for annotation of cancer is COSMIC [55].

The database is first implemented in 2004 [73] with four genes, such as HRAS, KRAS2, NRAS and BRAF. Since the effort of the COSMIC team the database is continuously updated with new finding and inclusion. In V91 (released on 07-APR-2020) has 577 curated genes of Tier-1. In V91 census data these four genes' Gene Symbol (Entrez GeneId; Role in Cancer) are HRAS (3265; ONGO), KRAS (3845; ONGO), NRAS (4893; ONGO) and BRAF (673; oncogene, fusion).

From 2018 onwards there is at least two releases for years; thus

To see the difference between updates of census (from 2018-2019-2020 Springs) the V84 of census Vs V88 of census Vs V91; Tier-1 ONGO, TSG, Fusion classes are compared as shown in Table.LA. From the Table. LA the highest (totally opposite function representation) change in the census; UBR5 is previously there in TSG class of V84; but in V88 and V91 it is ONGO.

**Table. LA:** Tracking the inclusion/removal/edition between Tier-1 ONGO, TSG and Fusion genes.

| Functionality | 2018(V84) | | 2019(V88) | | 2020(V91) | |
|---|---|---|---|---|---|---|
| | Gene Symbol | Entrez GeneId | Gene Symbol | Entrez GeneId | Gene Symbol | Entrez GeneId |
| ONGO | | | | | UBR5 | 51366 |
| TSG | UBR5 | 51366 | LATS1 | 9113 | | |
| | FAM46C | 54855 | LATS2 | 26524 | TENT5C | 54855 |
| | C2orf44 | 80304 | | | WDCP | 80304 |
| Fusion | ZNF198 | 7750 | | | ZMYM2 | 7750 |
| | NUTM2A | 728118 | | | NUTM2D | 728130 |

### 6.1.1 Data usage in project

The project and classification between ONGO, TSG and Fusion are based on the 2019 (V88) dataset. This data is used as (base data) used for model selection (between the performance of the models) for of the project/paper.

**Table. LB:** UBR5 gene's mapping information

| Gene Symbol (Entrez GeneId) | Cross reference of PDB | | | | | Primary structure length |
|---|---|---|---|---|---|---|
| UBR5(51366) | 1I2T | X-ray | 1.04 | A | 2393-2453 [»] | 60 |
| | 2QHO | X-ray | 1.85 | B/D/F/H | 180-230 [»] | 50 |
| | 3PT3 | X-ray | 1.97 | A/B | 2687-2799 [»] | 112 |

Since the Gene Symbol UBR5 (51366) have changed in opposite functionality. The impact of this negligible (from Table. LB) due to; from the results the PDBs with SITE information for classification performs better; thus, for the filtered (PDBs with primary structural length > threshold 81) data only 3PT3 satisfied and 3PT3 doesn't has SITE information. The models build based without SITE information affection due to this one PDB is negligible, since those data set (without) is around 2k range.

### 6.2 Refined(Cleaned) PDBs used changes due to the Uniprot mapping update

Since the PDBs are downloaded from Uniprot mapping; the Uniprot website itself update the mapping information. For an example, earlier Uniprot has "uni-gene" mapping; after the retirement of "uni-gene" that has removed as well in Uniprot. This shows the mapping website keep on updating itself (evolve with the new findings on the field).

#### 6.2.1 Impact due to mapping updates

Due to the recurrent census updates; to avoid/ develop the models in fast phase the mapped PDBs datasets always reused unless big change occurred. Thus, for V88 mapping of PDBs: the mapped PDBs from 2018 (V84) mapped on Spring -2018 were reused; only the new genes such as "LATS1" and "LATS2" PDBs are mapped and included.

In the deep learning models fully, cleaned dataset is used most. Means, after filtered (satisfy the threshold condition >81); The fully cleaned dataset means no overlap between the classes at all (like no overlap between neither ONGO and TSG, nor ONGO and Fusion nor, TSG and Fusion); used to train the deep learning models in this paper; thus this changes in mapping also have impact on the data selection for PDB (effect the final model used for classify). For an example,

GeneID "3020" is in both Census V84 and V91. Due to the cleaned dataset processing; I have found such one PDB id (5AY8); On earlier mapping (done around 2018 Fall) has the uniport mapping of PDB "5AY8" was there for GeneID 3020 (belong to T1 ONGO), but in 2020 Spring it's not mapped for GeneID 3020.

## 7. Conclusion

Effective preprocessing for 3D convolutional deep learning stage and some primary structure classification methods were proposed in this paper to classify the cancer genes: proto-oncogenes, Tumor suppressor genes and Fusion genes. Cancer progresses due to, proto-oncogenes mutation and become uncontrollable cell divisor, or cancer suppressor genes mutation and lose their function, or fusion genes formation effects the wanted functionality. By having a model that confidently identifies (with their corresponding probability; thus the user can decide the functionality the structure independently) proto-oncogene or cancer suppressor genes or fusion genes from the structure, this study proposing a new tool to discover a new set of cancer suppressor genes or proto-oncogenes or fusion (belongs to OG or TSG) genes that may not have been identified in the literature of having such functionality. By activating the quiescent potential TSG gene through drugs; the activated TSG genes superintends in controlling tumor growth. Indeed identified potential TSG genes have to be verified in wet lab models which mimic human gene content.

As per our knowledge this is the first study proposing classifying Fusion class along with OG and TSG. As per my knowledge this may first paper proposing methods such that using structural predictions combined with primary structural sequence information to predict the primary structural function prediction.

To compare the performance with the state-of-the-art models (which is developed to classify OG/TSG), the model's OG/TSG performance (AUROC) is compared as shown in the Table IV.II of Section 4.3. Where, the approach 1 model's classification of genes, through the methods' AUROC are above 0.94. Which outperforms all the previously developed methods (studies in this area; 0.924 is the highest AUROC reported till now as mentioned in section 4.3). And the approach 1's Ensemble methods' AUROC is 0.99, which outperforms the state-of-the-art models classifying OG/TSG to identify the cancer genes. Further, this studies' approach 1's models' DCNN AUROC is 0.978 which outperforms the [42]'s DCNN model. The study's approach 1's and approach 2's preprocessed three (OG/TSG/Fusion) class 3D-structure (PDB) validated and tested DCNNs' accuracy rate is 93.5% and 70.66% as shown in Section 4.1.1 and 4.2.1 repectively.

Further, the model applied on Tier-2's ONGO, and TSG (these Tier-2 genes' annotated classes like ONGO, TSG, etc are based on resent study's or studies' evidence towards the class or classes; may be those genes are potential ONGO or TSG as annotated). Tier-2's ONGO primary structures' preprocessed PDBs' DCNN results strengthen ~70% of them (those PDBs classified as ONGO); Likewise, in TSG, preprocessed PDBs' the DCNN results strengthen ~51% (classified as TSG). Further using this approach 1 DCNN results with methods in Tier-2 ONGO primary structures, obtained results strengthen five out of eight classified as ONGO; especially three of them have shown more supportive evidence towards the class since these three primary structures are classified by multiple PDB structures. Likewise, among ten out of fourteen TSG primary structures classified as TSG; especially seven of them have shown more supportive evidence towards the class since these seven primary structures are classified by multiple PDB structures.

The success of our model warrants our future study to apply the same deep learning model to humans' (GRCh38) genes, for predicting their corresponding probabilities of functionality in the cancer drivers. This may lead to another Tier-2 annotated genes for verification.

# Appendix-1: Sequence length threshold selection

**TABLE A1:** NUMBER OF PDB IDS WITH THE SEQUENCE OVERLAPPING LENGTH

| Bin | Frequency | cumulative frequency | Fraction covered by cumulative |
|---|---|---|---|
| 1 | 0 | 0 | 0.00 |
| 21 | 494 | 0 | 0.00 |
| 41 | 66 | 560 | 15.52 |
| 61 | 26 | 586 | 16.24 |
| 81 | 34 | 620 | 17.18 |
| 162 | 1225 | 1845 | 51.12 |
| 243 | 622 | 2467 | 68.36 |
| 324 | 452 | 2919 | 80.88 |
| 405 | 405 | 3324 | 92.10 |
| 486 | 72 | 3396 | 94.10 |
| 567 | 60 | 3456 | 95.76 |
| 648 | 73 | 3529 | 97.78 |
| 729 | 8 | 3537 | 98.00 |
| 810 | 2 | 3539 | 98.06 |
| 891 | 0 | 3539 | 98.06 |
| 972 | 14 | 3553 | 98.45 |
| 1053 | 11 | 3564 | 98.75 |
| 1134 | 36 | 3600 | 99.75 |
| 1215 | 8 | 3608 | 99.97 |
| More | 1 | 3609 | 100.00 |

The **TABLE. A1's**

**Bin:** represents the threshold sequence lengths for PDB_ids
**Frequency:** Number of PDB_ids has sequence length between the threshold length of previous **Bin** and the current **Bin**.
**Cumulative frequency:** Number of PDB_ids has sequence length up to the **Bin** given.

**Fraction covered by cumulative:** Up to what percentage of the all PDB_ids covered by the threshold length (given by the **Bin**).

For example, take the row "**Bin**" 81, which is highlighted in the table above. The PDB_id's overlapping sequence length (with the corresponding Gene_Symbol) has threshold up to 81. "**Frequency**" section has 34 PDB_ids, those PDB_ids (34) has overlapping sequence length between 61 <= seq_length < 81. "**Cumulative frequency**" section has 620 PDB_ids. Those PDB_ids' (620) overlapping sequence lengths were lower than 81 (seq_length < 81). "**Fraction covered by cumulative**" section has 17.18% fraction of PDB_ids. Those PDB_ids' fraction (17.18%) was obtained by **cumulative frequency/ Sum of PDBs** that means (100*620/3609 = 17.18%)

Already filtered PDB files (PDB files obtained by X-Ray diffraction) were only considered here. As mentioned in section 1 (the mapped data from UniProt [65], has Gene_Symbols with PDB ids and their corresponding, start-end overlap sequence with the gene). From that overlapping sequence length for each PDB file with their genes was gained by negating the starting position of the PDB file from ending position of the PDB file for coreesponding gene. Since some of the PDB files overlapping sequence length is too small, thus the contribution to the Gene from the isoform is less. To extract useful PDB files, the threshold length should be obtained (without loosing too many PDB files and reduce the noise created by small sequence length PDB files). If the PDB files have more than one sequence length, then those PDB files are left in the whole experiment.

To select the overlapping sequence threshold length, all the dataset (OG, TSG, and Fusion) with the sequence length) were combinedly considered. Table. A1 shows, how many PDB files fell between the overlapping sequence lengths (Bin). From the Table. A1, the threshold 81 chosen, because this make sure more than 82% PDB ids remained for further processing. Thus, it reduces the noise created by the PDB ids which has overlap sequence length<81.

## Appendix-2: Resolution factor selection for normalization

Since the PDB files (Isoforms) are gathered from different sources, the resolutions for each of them is also different. Hence, it must be normalized; however, the normalization reduces the positional information of the PDB files. Extracted coordinates of PDB files must be maintained at least some of the original positional information, thus it must be factored without much altering. To find the factor, the resolution of each chosen PDB files were pooled as shown in Table A2 and Fig A2 (obtained from Table A2; to visualize the resolution distribution). The significant factor as "2.25" is chosen, because the mean of the distribution is 2.258, the standard deviation is 0.566 and the most frequent bins are 2, and 2.5, and the mean of the selected bins is 2.25 ((1.5+2 +2.5 +3)/4 = 2.25). First, the PDB files coordinates are normalized (divided by using their corresponding resolution), then multiplied by 2.25 (this makes sure no positional information is loss due to normalization). In this way final < x, y, z > coordinates were found.

TABLE A2: FREQUENCY OF RESOLUTION OF PDB FILES FOR ONGO, TSG AND FUSION OF TIER.1

| Resolution Bin | ONGO Frequency | TSG Frequency | Fusion Frequency | All Frequency | general bin frequency percentage | Cumulative |
|---|---|---|---|---|---|---|
| 0.5 | 0 | 0 | 0 | 0 | 0.00% | 0.00% |
| 1 | 2 | 2 | 0 | 4 | 0.13% | 0.13% |
| 1.5 | 82 | 72 | 48 | 202 | 6.76% | 6.89% |
| 2 | 413 | 399 | 216 | 1028 | 34.39% | 41.28% |
| 2.5 | 351 | 373 | 120 | 844 | 28.24% | 62.63% |
| 3 | 281 | 257 | 87 | 625 | 20.91% | 90.43% |
| 3.5 | 104 | 104 | 29 | 237 | 7.93% | 98.36% |
| 4 | 18 | 16 | 4 | 38 | 1.27% | 99.63% |
| 4.5 | 6 | 2 | 1 | 9 | 0.30% | 99.93% |
| More | 1 | 0 | 1 | 2 | 0.07% | 100.00% |

These bins represent the occurrence of resolution in PDBs, eg: Bin 2 contains the resolution of PDB files between the count of 1.5 and 2 (1.5 < resolution of PDB file <= 2),

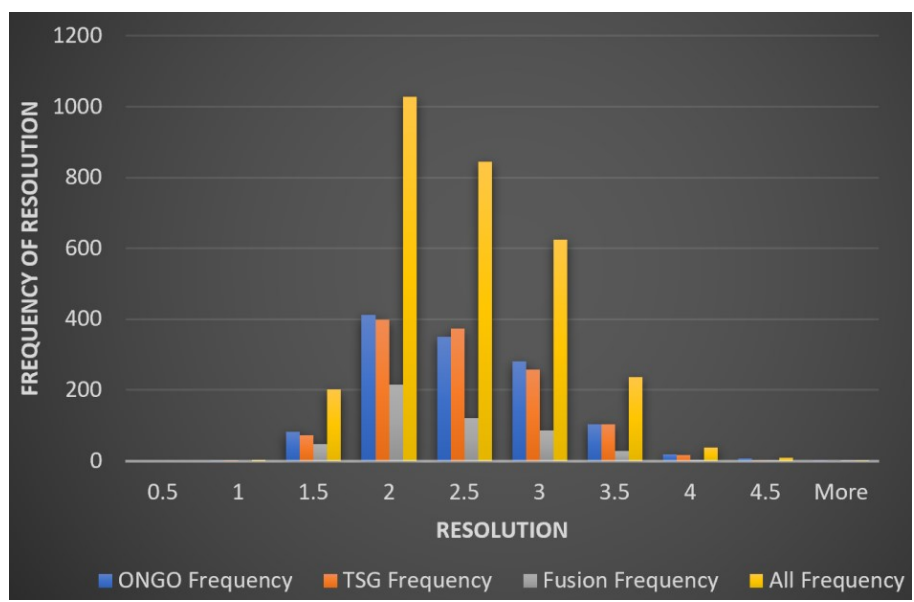

Fig. A2: Resolution of PDBs with their corresponding occurance(frequency)

Appendix-3: Surface Cα Indexing

The main piece of the experiment is surface atom detection. If proper surface atoms are not defined, then the inefficiency propagates through the deep learning model for predicting PDBs not efficiently. As per the work [42] try to develop an algorithm that determines the surface atoms (only considering Cα atoms) of the PDB. Instead of generating/improving surface detection algorithm better to use the already well-known advanced tool (MSMS [74]). To define surface atoms, there are newly developed tools as shown in [74] available, however MSMS is used widely. As mentioned in [75], "*MSMS allows computing very efficiently triangulations of Solvent Excluded Surfaces*".

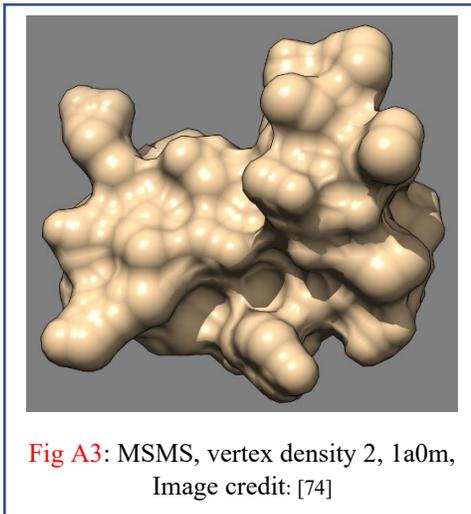

Fig A3: MSMS, vertex density 2, 1a0m, Image credit: [74]

The MSMS-tool only calculates the **solvent excluded** surface of the PDB-structure, but it doesn't explicitly mention the surface Cα atoms. Thus, in order to find the surface Cα atoms, two main ways considered with the tool such as,

1. The depth of each Cα atoms from the surface atoms: which is calculated by choosing the minimum distance between the surface and the Cα atoms, and then define that distance as the Cα atoms' depth from the surface.
2. The depth of each residue from the surface atoms: which is calculated by choosing the minimum distances between the surface and all atoms in the residue, and then average that distance as the residues' depth from the surface.

For the surface calculation, all the atoms in the PDB file are considered by MSMS tool. Afterward, for the depths calculation of the Cα atoms, the PDB files' (Chain ID assignment's) ATOM record was only considered, thus the (Chain ID assignment's) HETATM record was not included. Since, HETATM records are non-polymer structures thus, it doesn't have amino acid information (However they have Cα atoms, for our experiment the chemical properties allocated according to the amino acid details).

In order to define (or select) the depths threshold for selection of surface Cα atoms, only the PDB_ids(which satisfied up to SITE records) were considered. From that, in order to choose the threshold depth (the combination of depths, such as depth of Cα atoms and the depth of residue) for surface; two stats are mainly considered such as,

1. Check how much percentage of SITE groups' Cα atoms missed, when the threshold value is assigned.
2. Check how much percentage of Cα atoms chosen for the PDB while assigning that threshold.

The overall objective is choosing the threshold combination without losing approximately 10% SITE information and choose the important Cα only (which means the minimum number of Cα atoms as possible).

To select the threshold ranges (bins). First, all the SITE's depths were pooled. Then Cα's threshold bins selected from only considering the Cα threshold depth based on loss of the SITE structures between 7%-11%, and then the residue depth's threshold bin selection is produced separately as the same way as, depth of Cα atom bins.

For the threshold bins of Cα atoms depth selection; only the depths of Cα atoms from the surface, which has loss of atoms in SITE range [7%, 8%, 9%,10%,11%] were considered. From the experiment 7.2, 6.9, 6.7, 6.5, and 6.4 are chosen for the Cα threshold depth, which thresholds causes 7%, 8%, 9%,10%,and 11% loss of SITE atoms accordingly.

For the threshold bins of residue depth selection; only the depths of residue from the surface, which has loss of atoms in SITE range [7%, 8%, 9%,10%,11%] were considered. From the experiment 6.9, 6.7, 6.6, 6.5, and 6.3 are chosen for the residue threshold depth, which thresholds causes 7%, 8%, 9%,10%,and 11% loss of SITE atoms accordingly.

TABLE A3.I: MOTIF MISSING ATOMS %, WHILE VARYING THE THRESHOLDS OF DEPTHS

|  |  | Threshold of residue depth | | | | |
|---|---|---|---|---|---|---|
|  |  | 6.9 | 6.7 | 6.6 | 6.5 | 6.3 |
| Threshold of Cα depth | 7.2 | 9.10914 | 9.99487 | 10.5837 | 11.1871 | 12.5445 |
|  | 6.9 | 9.90982 | 10.6431 | 11.1373 | 11.6748 | 12.8863 |
|  | 6.7 | 10.6896 | 11.3106 | 11.7471 | 12.206 | 13.3019 |
|  | 6.5 | 11.7551 | 12.2493 | 12.6055 | 12.9826 | 13.9132 |
|  | 6.4 | 12.4258 | 12.8333 | 13.1334 | 13.4768 | 14.3144 |

After these bins were selected; both bins of thresholds were linked. Where the percentage of MOTIF atoms missed, was checked with the selection of surface using these threshold combinations. Thus, there are totally 25 combinations were checked as shown in the Table A3.I.

Threshold combinations as highlighted, the threshold depth of Cα as 7.2 and the threshold depth of residue as 6.7 is worked well as shown in Table A3.I. However, the highlighted portion thresholds were assigned separately to to find the percentage of Cα atoms chosen for surface.

The ratio of missing MOTIF atoms near to 10% and the percentage of Cα atoms missed is checked as shown in Table A3.II.

To choose the threshold, a factor is calculated as shown in equation (1). The highest factor's threshold combination was selected. The factor based on the missing percentage MOTIF surface atoms should be nearer to 10% and it must choose the minimum percentage of surface Cα atoms as possible.

$$\text{Thus, the factor} = \left| \frac{1}{(10 - \text{missed MOTIF atoms percentage}) \times \text{percentage of surface C}\alpha \text{ atoms}} \right| \quad -(1)$$

TABLE A3.II: PERCENTAGE OF Cα ATOMS CHOSEN FOR THRESHOLD CONDITIONS

|  |  |  | Residue threshold depth | | |
|---|---|---|---|---|---|
|  |  |  | 6.9 | 6.7 | 6.6 |
| ONCO | Cα threshold depth | 7.2 | 94.241 | 93.742 | 93.4545 |
|  |  | 6.9 | 93.7017 | 93.3184 | 93.1038 |
|  |  | 6.7 | 93.2101 | 92.8974 | 92.7255 |
| TSG | Cα threshold depth | 7.2 | 95.2399 | 94.8002 | 94.5313 |
|  |  | 6.9 | 94.7557 | 94.4162 | 94.193 |
|  |  | 6.7 | 94.3065 | 94.0331 | 93.8488 |
| Fusion | Cα threshold depth | 7.2 | 93.7676 | 93.0892 | 92.7362 |
|  |  | 6.9 | 93.3871 | 92.8099 | 92.4906 |
|  |  | 6.7 | 93.0233 | 92.5351 | 92.2521 |
| All | Cα threshold depth | 7.2 | 94.41617 | 93.87713 | 93.574 |
|  |  | 6.9 | 93.94817 | 93.51483 | 93.26247 |
|  |  | 6.7 | 93.5133 | 93.1552 | 92.94213 |

TABLE A3.III: FACTORS FOR FINAL THRESHOLD SELECTION

|  |  | Threshold of residue depth | | |
|---|---|---|---|---|
|  |  | 6.9 | 6.7 | 6.6 |
| Threshold of Cα depth | 7.2 | 0.011889 | 2.076456 | 0.01831 |
|  | 6.9 | 0.118032 | 0.01663 | 0.00943 |
|  | 6.7 | 0.01551 | 0.00819 | 0.00616 |

Factors' results are shown in Table A3.III. For an example the combination of threshold depth of Cα as 7.2, threshold depth of residue as 6.7's Factor = |1/ ((10-9.99487) x 93.87713) | = 2.076456

From the results, the combination of threshold depth of Cα as 7.2 and a threshold depth of residue as 6.7 were chosen. The surface Cα atom satisfy both depth threshold (Cα's depth from the surface less than these depths accordingly) chosen to calculate the final surface Cα coordinate < x, y, z > as mentioned in Appendix-2.

The progress given in the link:
https://drive.google.com/file/d/1h1AlUIadeGQLkpxfT_wlTLdPtawrVlcx/view?usp=sharing
supports the threshold selection of surface works. In the progress the all Cα used and the man threshold(which is less than the software threshold since only one man reported SITE information is available for selected ONGO and TSG class PDBs) are compared with seleted SOFTWARE threshold named as (named as SOFT threh in progress).

As expected SOFT thresh works, because the chosen factor is near to factor of ten; factor of ten widely used in engineering(e.g: power gain is expressed in decibels(dB) represent in factor of 10; in telecommunication power gain with Rule of 10s), as well as in biology(e.g:rule of 10 or 10% rule in energy pyramid, burns rule of tens etc.).

## Appendix-4: Base line model

Since the model is almost the same in [42]'s model. Thus as mentioned in in [42] the model consists of "*four convolution and pooling layers and three fully connected layers including the final classifier. Thus, the convolution kernel size (p), pooling strides ($s_i$), number of hidden neurons ($h_1$, $h_2$), convolution pad (γ), and the number of generated feature maps ($d_i$) are shown in*" Table A4. These parameters have been set up after several control experiments and initial evaluations. The CNN receives 24 projection's 128×128×21 feature maps in parallel and performs a multi-class (OG Vs TSG Vs Fusion) classification. Each layer is equipped by the Swish active function. This experiment used 50% dropout in the fully connected layers to control probably overtraining. The convolution/pooling layers extract 384×64 = 24,576 visual features for p=3. Since the model with p = 3 performed well if p=3 then the model has 2,533, 655 trainable parameters.

TABLE A4: THE BASE LINE CNN'S PARAMETERS.

| Parameter | Value | Parameter | Value |
|---|---|---|---|
| $d_1$ | 32 | $s_1$ | 4 |
| $d_2$ | 32 | $s_2$ | 2 |
| $d_3$ | 64 | $s_3$ | 2 |
| $d_4$ | 64 | $s_4$ | 2 |
| p | 3-7 | γ | 2 |
| $h_1$ | 100 | $h_2$ | 50 |

Table credit [42]

## Appendix-5: Calculation of Number of parameters

TABLE A5.: THE BRAIN INCEPTION RESIDUAL DCNN'S NUMBER OF TRAINABLE PARAMETER CALCULATION

| Layer wise CNN | 1-D CNN reducer | Single projection feature extractor (with CNN) | Applying same (1)/parallel feature extractors (Parallel: different models for each data given) | | Fully connected layer parameters calc | | | | | | Total fully connected parameters | Total parameters |
|---|---|---|---|---|---|---|---|---|---|---|---|---|
| | | | Number of NN | Total feature extractor param | Flatten | layer_1 | | layer_2 | | Softmax layer | | |
| | | | | | | neurons | Weights | neurons | Weights | | | |
| Layer_1 | | 32,512 | 1 | 32,512 | 6,144 | 100 | 614,500 | 50 | 5,050 | 153 | 619,703 | 2,925,751 |
| | 8,224 | | 48 | 394,752 | | | | | | | | |
| Layer_2 | | 85,248 | 1 | 85,248 | | | | | | | | |
| | 8,224 | | 48 | 394,752 | | | | | | | | |
| Layer_3 | | 85,248 | 1 | 85,248 | | | | | | | | |
| | 16,448 | | 48 | 789,504 | | | | | | | | |
| Layer_4 | | 129,280 | 1 | 129,280 | | | | | | | | |
| | 16,448 | | 24 | 394,752 | | | | | | | | |

The same number of parameters is brain inception without residual as well
**Proof** of parameter calculation: https://drive.google.com/open?id=1yCiy7a6UGdoriNw6mtXf-CXnbDV8doiO, https://drive.google.com/open?id=15AD41y8xLiTpUiMessWvrBPFMeItlySM, https://drive.google.com/open?id=1W1XkSRLdw-iOCwcUDabwyZpXejVwCyM, https://drive.google.com/open?id=1vzBLUukkr5_0bOFonAYGPvz11lY3SSwO

The inception residual convolution/pooling layers extract 24x2x2×64 = 6144 visual features. The model has 2,925,751 trainable parameters. And the parameter calculation is shown in Table A5.

For approach 2's

The total number of train data's feature = Number of PDBs x projections x size_x x size_y x channels(feature)

$$= 1628 \times 24 \times 128 \times 128 \times 21$$
$$= 1.34 \times 10^{10}$$

Ratio of Input features/Trainable parameter = $1.34 \times 10^{10} / 2,925,751 = 4604$

likewise, approach 1's total number of train data's feature is $1.56 \times 10^{10}$ and Ratio of Input features/Trainable parameter is 5,357.

## Appendix-6: Training and testing dataset selection from Tier 1

From the Table A6 just take the number of gene(primary structure) and PDBs details of 21- amino acid. The removed/cleaned data consists of 2108 PDB files belongs to 138 genes. From that, 1574 PDB files belong to 105 genes were used for training, and the remaining 33 genes and their corresponding 545 PDB files were used for testing as shown in Table A6. As mentioned in section 2, the training set's 1574 3D coordinate feature set was converted into three 2D feature sets (1574 x 24=37776). Thus, 37776 with 128 x 128 x 21 feature maps, were used to train the deep learning model.

TABLE A6: TRAINING AND TESTING DATASET SELECTION FROM TIER 1.

| Classes of Tier_1 | Data purpose | SITE (thresh>81) | | preprocessed | | cleaned preprocessed PDBs | |
|---|---|---|---|---|---|---|---|
| | | Number of Genes | Number of PDBs | Number of Genes | Number of PDBs | 21 amino acids (21-channels) | 20 amino acids (17-channels) |
| ONCO | train | N/A | N/A | 32 | 590 | 589 | 588 |
| | test | N/A | N/A | 13 | 448 | 445 | 445 |
| | Overall | 49 | 1121 | 45 | 1035 | 1031 | 1030 |
| TSG | train | N/A | N/A | 50 | 639 | 632 | 632 |
| | test | N/A | N/A | 14 | 79 | 79 | 79 |
| | Overall | 65 | 751 | 64 | 713 | 706 | 706 |
| Fusion | train | N/A | N/A | 23 | 403 | 353 | 353 |
| | test | N/A | N/A | 6 | 18 | 18 | 18 |
| | Overall | 30 | 425 | 29 | 421 | 371 | 371 |
| All | train | N/A | N/A | 105 | 1632 | 1574 | 1573 |
| | test | N/A | N/A | 33 | 545 | 542 | 542 |
| | Overall | 144 | 2212 | 138 | 2169 | 2108 | 2107 |

V88 gene PDB details are frozen; then used to train the models reported in this Table and progress (unless mentioned as V91 in name of the sheet)

Problematic (contain Overlapping PDBs) primary structures are evaluated separately, not in neither train nor test nor validation. Those primary structures' (are reported in Table I) and their corresponding PDBs (has SITE information and primary structural length>81) are evaluated separately. Thus, the column "SITE (thresh>81)" only reporting overall number of gene and PDB, the train and test are Not-Applicable (N/A).

Some the dataset numbers train, test, overall and clean different, due to these five reasons.

1. Training and Testing set separation is based on Primary structural separation, thus some Training and Testing PDBs has overlapping of eight PDBs. Here three of them belongs to ONGO and five of them belong to TSG.
2. In ONCO in training class 590 PDBs satisfied the condition but for the deep learning model 589 PDBs are used; since PDB file "4MDQ" not able produce results through MSMS tool.
3. In ONCO in testing class 448 PDBs satisfied the condition but the deep learning model classified only 445 PDBs. Among these three left the PDBs "3GT8" and "3LZB" has unknown amino acids so those left and PDB "721P" has no single $C\alpha$ surface atom by the threshold condition as mentioned in section Appendix-3.
4. In TSG "2H26" has unknown amino acids. PDBs "5C0B" and "5C0C" have size issues as their corresponding highest eigen vector's (x size) is 143 and 147 accordingly the feature size for 1 of the $8^{th}$ quadrants is 128. And 4 PDBs are in overlap, altogether 7 PDBs.
5. In Fusion, the 50 PDBs belongs to overlapping with other classes.

## Appendix-7: Tier_1 and Tier-2 results

Further Tier_1 and Tier_2 overall classification is attached in link (https://drive.google.com/drive/folders/1YCuMVPhAy7tIdGFiebInZ_6-kBAH1kMM?usp=sharing).

TABLE A7.I: ANNOTATION OF NOT-ANNOTATED GENES

| Probabilities of direct pikles | | | | |
|---|---|---|---|---|
| Uni_gene_ID | OG | TSG | Fusion | Class |
| 1500 | 0.46 | 0.28 | 0.26 | ONGO_TSG_Fusion |
| 1630 | 0.82 | 0.17 | 0.01 | ONGO |
| 5579 | 0.95 | 0.04 | 0.01 | ONGO |
| 6000 | 0.23 | 0.48 | 0.29 | TSG_Fusion |
| Probabilities from Ensamble | | | | |
| Uni_gene_ID | OG | TSG | Fusion | Class |
| 1558 | 0.89 | 0.06 | 0.05 | ONGO |
| 2042 | 0.72 | 0.19 | 0.09 | ONGO |
| 2045 | 0.44 | 0.5 | 0.06 | ONGO_TSG |
| 2316 | 0.65 | 0.29 | 0.06 | ONGO_TSG |
| 286 | 0.23 | 0.59 | 0.18 | TSG |
| 30835 | 0.42 | 0.5 | 0.08 | ONGO_TSG |
| 3685 | 0.31 | 0.66 | 0.04 | ONGO_TSG |

Inorder to show the results (in the link) representation by the model and methods. The Table A7.I presents the results of Not-annotated genes of Tier-2 (from COSMIC V91 [55] data) using V88 models' as explained above. Annotations of these predicted structures' classes assigned in different ways(depends user's/annotator's purpose) from the given probability.

For an example, take the gene-id 3685 (it's gene symbol is ITGAV) if the user just need the mainly contributing class then the geneID can be showing higher probability to TSG class as 0.66 ( which is >0.5), on the other hand if the user need multi classification then the ONGO can be included with TSG; since the ONGO probability is 0.31 (which > 0.25).

Thus, here the annotation is for muti classififaction. And the classes are annotated as shown in the psudocode Fig A7.1. There the

m_p_d: Minimum probability decision maker (assigned as 0.25)

P_ONGO: Predicted probability of ONGO class (e.g: for GeneID 3685 probability is 0.31)

P_TSG : Predicted probability of TSG class (e.g: for GeneID 3685 probability is 0.66)

P_Fusion: Predicted probability of Fusion class (e.g: for GeneID 3685 probability is 0.04)

The probabilities should be added to 1.0. Since the probabilities are rounded to $2^{nd}$ digit, due to three class sometimes these round up end with 1.01 or 0.99. From the example presented here if the probabilities are rounded to $3^{rd}$ digit 0.306, 0.656 and 0.038; then the addition is end up with 1.0.

```
if P_ONGO >= m_p_d and P_TSG >= m_p_d and P_Fusion >= m_p_d:
        return "ONGO_TSG_Fusion"
elif P_ONGO >= m_p_d and P_TSG >= m_p_d and P_Fusion < m_p_d:
        return "ONGO_TSG"
elif P_ONGO >= m_p_d and P_TSG < m_p_d and P_Fusion >= m_p_d:
        return "ONGO_Fusion"
elif P_ONGO < m_p_d and P_TSG >= m_p_d and P_Fusion >= m_p_d:
        return "TSG_Fusion"
elif P_ONGO >= m_p_d x 2:
        return "ONGO"
elif P_TSG >= m_p_d x 2:
        return "TSG"
elif P_Fusion >= m_p_d x 2:
        return "Fusion"
```

**Fig A7.1:** Pseudo code for annotating classes

**Tier-2 classification of ONGO, TSG and Fusion**

**Functional classification of left PDBs (Isoforms)**

Since Approach 2 performance is comparably low (due to lack of training data); these classifications results shown below are obtained by Approach 1. In those classification results if the PDB is not presented in none of Training set of 10-folds, then the classification of 10-models (deep learning) are used to ensemble the results.

**This can be done in two ways**

**Way-1:** Just consider the PDBs that were not presented in training the models. (Include new T_1 PDBs due to update). Fortunately, due to only few PDBs are included in the new- mapping (preprocessing). Those PDBs and their corresponding probability details is shown in Table A7.II.

And the classification class match with the given class as confusion matrix in Fig.A7.2.

| Given class | predicted class | | |
|---|---|---|---|
| | ONGO | TSG | Fusion |
| ONGO | 20 | 66 | 11 |
| TSG | 9 | 72 | 35 |
| Fusion | 0 | 6 | 1 |

**Fig. A7.2**: Confusion matrix of Way-1 predicted PDBs

TABLE A7.II: TIER-1 NEWLY INCLUDED PDBs' WITH PREDICTED PROBABILITIES

| Given class | PDBs | Ensembled predicted prob | | |
|---|---|---|---|---|
| | | ONGO | TSG | Fusion |
| TSG | 5B5W | 0.47 | 0.36 | 0.17 |
| TSG | 5BRK | 0.04 | 0.75 | 0.21 |
| Fusion | 5AY8 | 0.4 | 0.16 | 0.45 |

**Way-2:** Only Tier-2 classes' PDBs were considered. And the classification class match with the given class as confusion matrix in Fig. A7.3.

| Given class | predicted class | | |
|---|---|---|---|
| | ONGO | TSG | Fusion |
| ONGO | 20 | 65 | 11 |
| TSG | 9 | 71 | 35 |
| Fusion | 0 | 6 | 0 |

**Fig. A7.3**: Confusion matrix of Way-2 predicted PDBs

TABLE A7.III: PREDICTED AVERAGE PROBABILITIES OF PDBS WITH GIVEN CLASS, WAY-1 AND WAY-2.

| Way | Given class | Ensembled predicted probabilities average | | | number of PDBs |
|---|---|---|---|---|---|
| | | ONGO | TSG | Fusion | |
| Way-1 | ONGO | 0.592 | 0.353 | 0.054 | 29 |
| | TSG | 0.410 | 0.467 | 0.123 | 144 |
| | Fusion | 0.318 | 0.556 | 0.126 | 47 |
| Way-2 | ONGO | 0.592 | 0.353 | 0.054 | 29 |
| | TSG | 0.412 | 0.465 | 0.122 | 142 |
| | Fusion | 0.316 | 0.565 | 0.119 | 46 |

Further, the average of the whole newly include PDBs' probability is shown in Table A7.III (Note: this is not related to confusion matrix; the confusion matrix only represents the predicted classes).

**Note**: some of the PDBs are skipped due to not supporting the i/p-format of the model or contain lack of information (like not-annotated amino acids)

In Fusion class 6DEC skipped due to preprocessing issue; that has unknown amino acid.

In Fusion class PDBs 1O04, 1ZUM, and 3N80 optimally tilted direction has x size: 151, 144, and 142, thus those (6DEC, 1O04, 1ZUM, and 3N80) PDBs skipped in this experiment.

**Summary of Tier-2 PDB structure classification results:**

In Tier-2, among 29 ONGO, PDBs, the model results strengthen twenty of them (classified as ONGO); Likewise, among 142 TSG, PDBs the model results strengthen 72 (classified as TSG).

**Classification of primary structure of Tier-2 of V91**

As explained earlier only the PDBs satisfy the preprocessing condition to pass through the deep learning models are used to calculate the weightage of the probability of primary structure. The confusion matrixes of predicted classes with given methods is shown in Fig A7.4 to Fig A7.8; further binary classification of ONGO Vs TSG confusion matrixes with different methods are shown in Fig A7.9 to Fig A7.13.

Further, the average probabilities of given class with predicted probabilities is shown in Table A7.IV(A) and Table A7.IV(B); as mentioned in earlier section these table just represent the average of probabilities not related to classification.

From the confusion matrixes the Fig A7.3-Fig A7.7 and binary-classifications Fig.8 to Fig.12 both shows same count since the classified class as Fusion class is zero.

TABLE A7.IV(A): PRIMARY STRUCTURES PREDICTED AVERAGE PROBABILITIES OF GIVEN CLASS

| Methods classes | Predicted mean probability | | | | | |
|---|---|---|---|---|---|---|
| | Direct | | | Ensemble | | |
| | ONGO | TSG | Fusion | ONGO | TSG | Fusion |
| ONGO | 0.31 | 0.63 | 0.06 | 0.67 | 0.27 | 0.06 |
| TSG | 0.25 | 0.69 | 0.06 | 0.39 | 0.46 | 0.15 |
| Fusion | 0.39 | 0.57 | 0.04 | 0.36 | 0.56 | 0.08 |

| Given class | predicted class | | |
|---|---|---|---|
| | ONGO | TSG | Fusion |
| ONGO | 2 | 1 | 1 |
| TSG | 2 | 3 | 2 |
| Fusion | 0 | 0 | 0 |

**Fig A7.4**. Confusion matrix of Direct

TABLE A7.IV(B): PRIMARY STRUCTURES PREDICTED AVERAGE PROBABILITIES OF GIVEN CLASS

| Methods classes | Predicted mean probability | | | | | | | | |
|---|---|---|---|---|---|---|---|---|---|
| | Method_1 | | | Method_2 | | | Method_3 | | |
| | ONGO | TSG | Fusion | ONGO | TSG | Fusion | ONGO | TSG | Fusion |
| ONGO | 0.66 | 0.28 | 0.06 | 0.72 | 0.21 | 0.07 | 0.62 | 0.31 | 0.07 |
| TSG | 0.40 | 0.45 | 0.14 | 0.42 | 0.47 | 0.12 | 0.36 | 0.46 | 0.18 |
| Fusion | 0.35 | 0.56 | 0.09 | 0.39 | 0.54 | 0.07 | 0.35 | 0.58 | 0.07 |

**Summary of Tier-2 primary structure results:**

In Tier-2, among eight ONGO primary structures, the model + method results strengthen five of the findings (classified as ONGO); especially three of them have shown more supportive evidence towards the class since these Three primary structures are classified by multiple PDB structures. Likewise, among fourteen TSG primary structures the model + method results strengthen ten of the findings (classified as TSG); especially seven of them have shown more supportive evidence towards the class since these seven primary structures are classified by multiple PDB structures.

| Given class | predicted class | | |
|---|---|---|---|
| | ONGO | TSG | Fusion |
| ONGO | 3 | 3 | 2 |
| TSG | 1 | 7 | 4 |
| Fusion | 0 | 0 | 0 |

**Fig A7.5.** Confusion matrix of ensemble

| Given class | predicted class | | |
|---|---|---|---|
| | ONGO | TSG | Fusion |
| ONGO | 4 | 4 | 1 |
| TSG | 0 | 6 | 5 |
| Fusion | 0 | 0 | 0 |

**Fig A7.6.** Confusion matrix of Method-1

| Given class | predicted class | | |
|---|---|---|---|
| | ONGO | TSG | Fusion |
| ONGO | 3 | 4 | 2 |
| TSG | 1 | 6 | 4 |
| Fusion | 0 | 0 | 0 |

**Fig A7.7.** Confusion matrix of Method-2

| Given class | predicted class | | |
|---|---|---|---|
| | ONGO | TSG | Fusion |
| ONGO | 3 | 4 | 2 |
| TSG | 1 | 6 | 4 |
| Fusion | 0 | 0 | 0 |

**Fig A7.8.** Confusion matrix of Method-3

| Given class | predicted class | |
|---|---|---|
| | ONGO | TSG |
| ONGO | 2 | 1 |
| TSG | 2 | 3 |

**Fig A7.9.** ONGO Vs TSG Confusion matrix of Direct

| Given class | predicted class | |
|---|---|---|
| | ONGO | TSG |
| ONGO | 3 | 3 |
| TSG | 1 | 7 |

**Fig A7.10.** ONGO Vs TSG Confusion matrix of ensemble

| Given class | predicted class | |
|---|---|---|
| | ONGO | TSG |
| ONGO | 4 | 4 |
| TSG | 0 | 6 |

**Fig A7.11.** ONGO Vs TSG Confusion matrix of Method-1

| Given class | predicted class | |
|---|---|---|
| | ONGO | TSG |
| ONGO | 3 | 4 |
| TSG | 1 | 6 |

**Fig A7.12.** ONGO Vs TSG Confusion matrix of Method-2

| Given class | predicted class | |
|---|---|---|
| | ONGO | TSG |
| ONGO | 3 | 4 |
| TSG | 1 | 6 |

**Fig A7.13.** ONGO Vs TSG Confusion matrix of Method-3

# Appendix-8 Brain Inception residual overall architecture

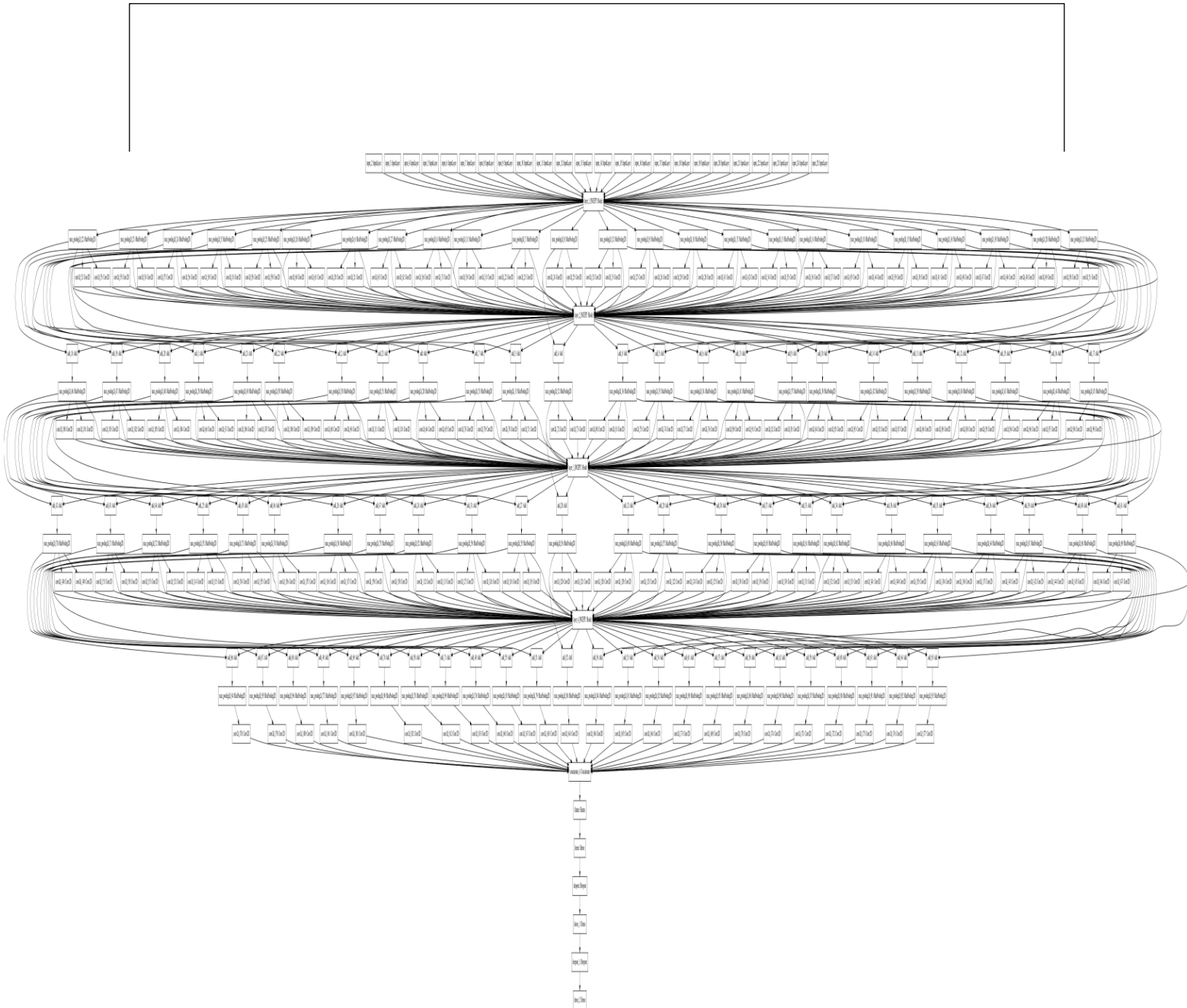

Fig A8: Brain inception residual full architecture

# Appendix-9: Example for methods using hypothetical sequencial data

Hypothetical example of Method-1 is shown below

[long---------------------- overlapped_ part -------]

It find the longest overlapped PDB and then use that to give the fractions for the rest of the PDBs

```
Gene   [150--------------------------------------------------------------------------------]
PDB_1            [60-------------------------------]
PDB_2      [45--------------20------]
PDB_3              [30--------30----------]
PDB_4                              [55----15-------------]
PDB_5                                                      [----12-----]
```

This algorithm first chooses the PDB_1 because that is the longest among all
Then calculate the fraction of the rest of the PDBs overlapped with that PDB_1

  PDB_1: 1
  PDB_2: 20/60 = 0.333
  PDB_3: 30/60 = 0.5
  PDB_4: 15/60 = 0.25
  PDB_5: 0

    with these fractions the length considered also saved to calculate the overall probability
here the length is PDB_1's that is 60

Then calculate the remaining length after removing the overlapped part
    PDB_1
PDB_2      [25----------]
PDB_3
PDB_4                              [40-------------]
PDB_5                                                      [----12-----]

This will lead to go to PDB_4 because that has the highest length[40]
Then follow the same procedure to calculate the fractions

PDB_1: 0
PDB_2: 0
PDB_3: 0
PDB_4: 1
PDB_5: 0

    with these fractions the length considered also saved to calculate the overall probability
here the length is PDB_1's that is PDB_4's: 40
after done every calculation:

$$\frac{60\ (P[PDB\_1] + 0.33\ P[PDB\_2] + 0.5\ P[PDB\_3] + 0.25\ P[PDB\_4])}{[40 + 60 + ...\ ]} + \frac{40\ (P[PDB\_4])}{[40 + 60 + ...\ ]}$$

Hypothetical example of Method-2 and Method_3 is shown below
Method_2 and Method_3 is mainly focused on covering the maximum gene as possible
  with the PDBids of that gene but how we choose those PDBs
       Method_2 uses non-overlapping PDB's to cover the gene
       Method_3 uses overlapping PDB's to cover the gene
     For ex:
  Gene    [150-----------------------------------------------------------------------]
  PDB_1              [60------------------------------]
  PDB_2     [45---------------20------]
  PDB_3     [45---------------20------]
  PDB_4                [30--------30----------]
  PDB_5                [10----10----]
  PDB_6                                     [55----15--------------]
  PDB_7                                                            [----12-----]

  Both algorithms uses the unique starting point for caculation
  Method_2 uses unoverlapping PDB's to cover the gene
  cov(0) = 0
  cov(1) = 45 PDB_2 and PDB_3 (finally we use this information to give the vote for that part by the probabilities)
  cov(2) = 60 PDB_1
  cov(3) = 60 PDB_1 when checking PDB_4
  cov(4) = 60 PDB_1 when checking PDB_5
  cov(5) = 55+45 = 100 ((PDB_2 and PDB_3) and PDB_6)
  cov(6) = 112 ((PDB_2 and PDB_3),PDB_6 and PDB_7)

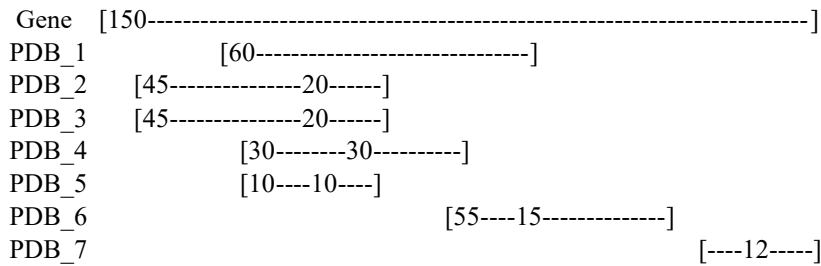

$$\text{Method\_2\_fin} = \frac{45\,(0.5\,P[PDB\_2] + 0.5\,P[PDB\_3]) + 55\,P[PDB\_6] + 12\,P[PDB\_7]}{112 \qquad\qquad 112 \qquad\qquad 112}$$

Method_3 uses overlapping PDB's to cover the gene

  cov(0) = 0
  cov(1) = 45 PDB_2 and PDB_3 (finally we use this information to give the vote for that part by the probabilities)
  cov(2) = 85 (PDB_2 and PDB_3) and PDB_1
  cov(3) = 85 (PDB_2 and PDB_3) and PDB_1 when checking PDB_4
  cov(4) = 85 (PDB_2 and PDB_3) and PDB_1 when checking PDB_5
  cov(5) = 125 ((PDB_2 and PDB_3), and PDB_6)
  cov(6) = 137 ((PDB_2 and PDB_3), PDB_6 and PDB_7)

Method_3_fin =
$$\frac{(45-10/2)(0.5\,P[PDB\_2] + 0.5\,P[PDB\_3]) + (40-15/2)P[PDB\_6] + 12\,P[PDB\_7]}{137}$$

# Appendix-10: Complexity of the methods

To calculate the complexity, for hypotheticalmodel assume a primary sequence contain maximum "n" number of PDBs. Since all these methods use same kind of data structure to do the calculation. The complexity of the common calculation is given below.

Complexity of running time is given in Font colour red
Complexity of memory is given in Font colour green

**Complextity of Method-1**

Pseudo code representation 2 for Method_1
1. **while** high_length > 0: O(n) or Ω(1)
2.     high_length, sel_PDB_id =select_pdb_with_high_length(un_overlap) O(n) and O(n)
3.     **if** high_length == 0:Θ(1)
4.         **break** Θ(1)
5.     un_overlap, over_lapped_sat = method_1_find_over_lab_with_sel(sel_PDB_id, un_overlap, high_length) O(n)

In pseudo code representation 2 for Method_1
line 1: if none of the PDBs' primary structure is overlapped with each then the while loop iterate " n" times thus upper bound at O(n). If highest length primary structure PDB is overlapped with all the rest of the PDBs then while loop end in first iteration so it's lower bound is Ω(1) line 2: function "select_pdb_with_high_length" finds the PDBs' primary structures highest length among them it goes through the unoverllaped PDBs(mean it not overllaped part with the previous highest length PDB's primary structure). It need maximum time is "n" go through all, and space to store the length thus higher bound is O(n) for both. And the number of unoverlap PDBs is reducing over each run.

line 5: "method_1_find_over_lab_with_sel" function find the fractions of overlapped primary structures with the highest length as "over_lapped_sat" thus it need to go through the all the unoverlapped PDBs to find the fractions thus higher bound is O(n) for both.

So the time complexity of method-1 is $O(n^2)$ or Ω(n) and the space complexity is O(n).

**Complextity of Method-2 and Method_3**

In pseudo code for Method_2 and Method_3
Starting positions of PDB_ids (for the group) are sorted in ascending order: that only taking sorting complexity like O(nlog n) can be achieved by quick sort, but here two or more PDB_ids has same start end position both it considered as one for the calculation thus, PDBs primary structures with same starting position and end position calculation need $O(n^2)$ time complexity, and space O(n) complexity.

Both **Method-2** and **Method_3**, need doubly nested loop structure to perform the bottom up dynamic programming approach, thus it need $\Theta(n^2)$ time complexity. Since both needed to store the indexes of each values of steps to go back and find out so the space complexity is always upper bounded by $O(n^2)$.

But after the calculation both needed to check the overlapped portion; that needs $O(n^3)$ time complexity; since it have to go through each groups created by buttom up fashion and go through all PDBs' primary structures to assign weights depends on do these PDBs' primary structures overlap or not(that checking need another loop). Space complexcity remains same.

Thus Method_2's and Method_3's time complexity is $O(n^3)$ and the space complexity is $O(n^2)$.